\pdfoutput=1

\documentclass[11pt]{article}

\usepackage{acl}

\usepackage{times}
\usepackage{latexsym}
\usepackage{CJKutf8}

\usepackage{graphicx}
\usepackage{subcaption,booktabs}
\usepackage{tabularx}
\usepackage{multirow, amsmath}
\usepackage{placeins}
\usepackage{gensymb} 
\usepackage{amssymb}

\usepackage[T1]{fontenc}

\usepackage[utf8]{inputenc}

\usepackage{microtype}

\title{\textsc{GeoMLama}: Geo-Diverse Commonsense Probing on Multilingual Pre-Trained Language Models}

\author{
	Da Yin 
\quad	Hritik Bansal
\quad	Masoud Monajatipoor 
\quad	Liunian Harold Li
\quad	Kai-Wei Chang \\
	Computer Science Department, University of California, Los Angeles\\
	{\tt \{da.yin,hbansal,liunian.harold.li,kwchang\}@cs.ucla.edu,}
	\\  
	{\tt monajati@ucla.edu} \\
}

\begin{document}
\maketitle
\begin{abstract}
Recent work has shown that Pre-trained Language Models (PLMs) store the relational knowledge learned from data and utilize it for performing downstream tasks. However, commonsense knowledge across different regions may vary. For instance, the color of bridal dress is \textit{white} in \textit{American} weddings whereas it is \textit{red} in \textit{Chinese} weddings.
In this paper, we introduce a benchmark dataset, \textbf{Geo}-diverse Commonsense \textbf{M}ultilingual \textbf{La}nguage \textbf{M}odels \textbf{A}nalysis (\textsc{GeoMLama}), for probing the diversity of the relational knowledge in multilingual PLMs.
\textsc{GeoMLama} contains 3,125 prompts in English, Chinese, Hindi, Persian, and Swahili, with a wide coverage of concepts shared by people from American, Chinese, Indian, Iranian and Kenyan cultures. We benchmark 11 standard multilingual PLMs on \textsc{GeoMLama}. Interestingly, we find that 1) larger multilingual PLMs variants do not necessarily store geo-diverse concepts better than its smaller variant; 2) multilingual PLMs are not intrinsically biased towards knowledge from the Western countries (the United States); 3) the native language of a country may not be the best language to probe its knowledge and 4) a language may better probe knowledge about a non-native country than its native country. Code and data are released at \url{https://github.com/WadeYin9712/GeoMLAMA}.
\end{abstract} 

\section{Introduction}
Pre-trained Language Models (PLMs)~\cite{peters-etal-2018-deep,radford2019language,devlin-etal-2019-bert,brown2020language} are increasingly used in various Natural Language Processing (NLP) applications.
Pre-trained on large-scale text corpora, they are shown to store relational knowledge~\cite{petroni-etal-2019-language,jiang-etal-2020-know,kassner-etal-2021-multilingual}, e.g., commonsense knowledge~\cite{zhou2020evaluating,lin-etal-2020-birds,nguyen2021advanced,zhou-etal-2021-rica}. 
They have been used to construct knowledge bases while requiring limited human effort for rule creation and validation~\cite{bosselut-etal-2019-comet,zhou-etal-2022-prix}. 

\begin{figure}[t]
\centering
\includegraphics[width=\columnwidth, trim=0 10 0 0, clip]{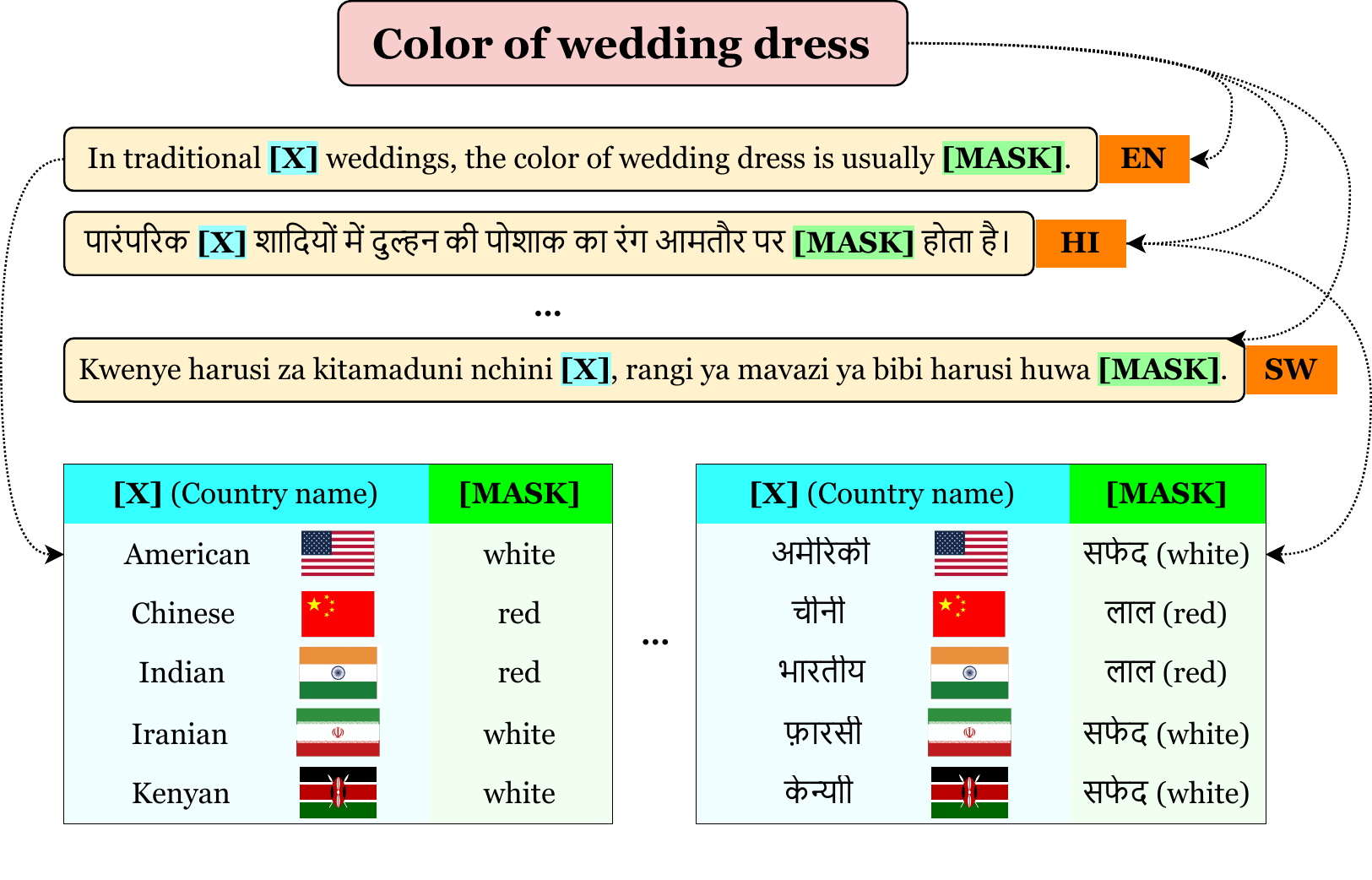}
\caption{Examples of prompts and gold answers in \textsc{GeoMLama}. For each concept (e.g., color of wedding dress), there are multiple masked multilingual prompts (\textcolor{orange}{English}, \textcolor{orange}{Hindi}, \textcolor{orange}{Swahili}, etc.) with specified country information \colorbox[rgb]{ .2,  1,  1}{[X]} querying geo-diverse knowledge about the concept. We test multilingual PLMs by examining the extent to which masked word predictions align with the gold answers in \colorbox[rgb]{ 0,  1,  0}{[MASK]} columns.}
\label{fig-intro}
\end{figure}

However, \emph{do PLMs store geo-diverse commonsense knowledge?} Geo-diverse commonsense~\cite{yin-etal-2021-broaden} is a collection of commonsense locally shared by people from certain regions but may not apply in other regions due to cultural and geographic differences. For instance, the color of bridal outfit in American wedding is white, while it is normally red in traditional Chinese and Indian weddings. PLMs which are unaware of geo-diverse knowledge may have disparity in performance on test data associated with different regions. This may lead to disadvantage of users in certain regions and further amplify bias in AI applications, such as constructing Western-centric knowledge bases eventually. 

In this paper, we concentrate on evaluating \emph{multilingual} PLMs~\cite{devlin-etal-2019-bert,conneau2019cross,conneau-etal-2020-unsupervised}. 
Studying geo-diversity naturally involves multilinguality. People in different regions may speak different languages, and it is natural to assume that geo-specific knowledge is better represented in its native language. Moreover, pre-trained on a collection of multilingual corpora, multilingual PLMs accumulate the knowledge from various languages. Therefore, we posit that knowledge in multilingual PLMs is more diverse than that in models trained on a single language. 

Centered around multilingual PLMs, we follow the original knowledge probing task LAnguage Model Analysis (\textsc{Lama})~\cite{petroni-etal-2019-language} and  introduce a new \emph{geo-diverse} probing benchmark \textsc{GeoMLama}. As shown in Figure \ref{fig-intro}, given a masked geo-diverse prompt with a particular country name \colorbox[rgb]{ .2,  1,  1}{[X]}, such as ``\textit{In traditional} \colorbox[rgb]{ .2,  1,  1}{[X]} \textit{weddings, the color of wedding dress is usually} \colorbox[rgb]{ 0,  1,  0}{[MASK]}.'', and a corresponding candidate answer list, \{``\textit{red}'', ``\textit{white}'', ``\textit{black}'', ``\textit{blue}'', ...\}, multilingual PLMs are required to predict the masked word \colorbox[rgb]{ 0,  1,  0}{[MASK]} from the candidate list. 

The characteristics of \textsc{GeoMLama} are summarized as follows. 1) \emph{Diverse answers across countries}: Each prompt is designed based on geo-diverse concept (e.g., color of traditional wedding dress in Figure~\ref{fig-intro}) and gold answers for masked word are different across countries. 2) \emph{Broad coverage of geo-diverse concepts}: \textsc{GeoMLama} encompasses comprehensive geo-diverse topics including habits and personal choices, cultures and customs, policies and regulations, and geography. 3) \emph{Coverage of multiple countries and languages}: \textsc{GeoMLama} involves knowledge about the United States, China, India, Iran, and Kenya, 
and is constructed by the native languages of the five countries, English, Chinese, Hindi, Persian, and Swahili.
Overall, there are 3,125 prompts in our benchmark.

We perform in-depth probing analysis on 11 multilingual PLMs, including mBERT~\cite{devlin-etal-2019-bert}, XLM~\cite{conneau2019cross}, XLM-R~\cite{conneau-etal-2020-unsupervised}, mT5~\cite{xue-etal-2021-mt5}, and XGLM~\cite{lin2021few}. In general, we observe that multilingual PLMs significantly outperform random guess, suggesting that multilingual PLMs are capable of storing geo-diverse commonsense to some extent. We then conduct fine-grained investigation across three dimensions. 

We first study the correlation between model performance and \emph{model size}. Contrary to our intuition, we notice that the largest models do not necessarily have the best performance on our benchmark. We further study \emph{the best language to probe the knowledge about a particular country}. Surprisingly, we find that the best language is not the native language of the given country (e.g., English is not the best language to probe knowledge about the US). We also explore \emph{the knowledge that can be most accurately probed by a particular language}. Similarly, we find that the most accurately probed knowledge is not the one about indigenous country of the language (e.g., the country for which Chinese prompts provide the most accurate predictions is not always China). Lastly, we find evidence of reporting bias that might explain such observations.

\section{Related Works}
\paragraph{Knowledge Probing on PLMs.}
\citet{petroni-etal-2019-language} first explore whether PLMs have capacity of storing factual knowledge about entities. 
Based on this observation, prior works involving knowledge probing focus primarily on creating more effective probing methods to elicit factual knowledge~\cite{jiang-etal-2020-know,jiang-etal-2020-x,shin-etal-2020-autoprompt,zhong-etal-2021-factual} or analyzing whether other types of knowledge are stored in PLMs~\cite{talmor-etal-2020-olmpics,zhou2020evaluating,kassner-etal-2021-multilingual,sung-etal-2021-language}. In the second line of works, there is a great variety of commonsense knowledge being explored, including social~\cite{zhou2020evaluating}, numerical~\cite{lin-etal-2020-birds} and spatial~\cite{zhang-etal-2020-language-embeddings,liu-etal-2022-things} commonsense. \textsc{GeoMLama} focuses on probing a new commonsense type, geo-diverse commonsense, on multilingual PLMs.

\paragraph{Multilingual Knowledge Probing and Multilingual Commonsense.}
MLAMA~\cite{kassner-etal-2021-multilingual} and Prix-LM~\cite{zhou-etal-2022-prix} simply focus on capturing multilingual factual knowledge about entities. XCOPA~\cite{ponti-etal-2020-xcopa} and X-CSR~\cite{lin-etal-2021-common} are two multilingual commonsense benchmarks, but both are built by translation from English commonsense benchmarks, without any consideration of region-specific commonsense. Different from prior works, we value geo-diversity and quantify the extent to which multilingual PLMs master such geo-diverse commonsense.

\paragraph{Geo-Diverse Commonsense.}
Geo-diverse commonsense is strongly correlated with cultures and geographic locations. There have emerged a few works~\cite{Acharya2020AnAO,yin-etal-2021-broaden,liu-etal-2021-visually,shwartz-2022-good} studying geo-diverse commonsense. Specifically, by collecting responses to questionnaire, \citet{Acharya2020AnAO} analyze the cultural difference between US and India about scenarios including wedding and funeral. \citet{yin-etal-2021-broaden,liu-etal-2021-visually} propose geo-diverse multimodal benchmarks, GD-VCR and MaRVL. They find that due to lack of geo-diverse knowledge, large performance disparity appears when multimodal models are applied on tasks requiring knowledge about Western and non-Western regions. \citet{shwartz-2022-good} propose culture-specific time expression grounding task to acquire specific temporal commonsense in different countries from multilingual corpora and models.

\paragraph{Inclusion in NLP.} Enhancing inclusivity of language processing technology and ensuring it works for everyone is essential. Several studies have focused on improving language inclusion~\cite{joshi-etal-2020-state,faisal-etal-2022-dataset}, gender inclusion~\cite{cao-daume-iii-2021-toward,dev-etal-2021-harms,lauscher-etal-2022-welcome}, and race inclusion~\cite{field-etal-2021-survey}. We hope that \textsc{GeoMLama} can enable future development in improving the diversity of knowledge embedded in pre-trained language models. 

\begin{figure*}[t]
\centering
\includegraphics[width=0.87 \textwidth, trim=0 145 10 0, clip]{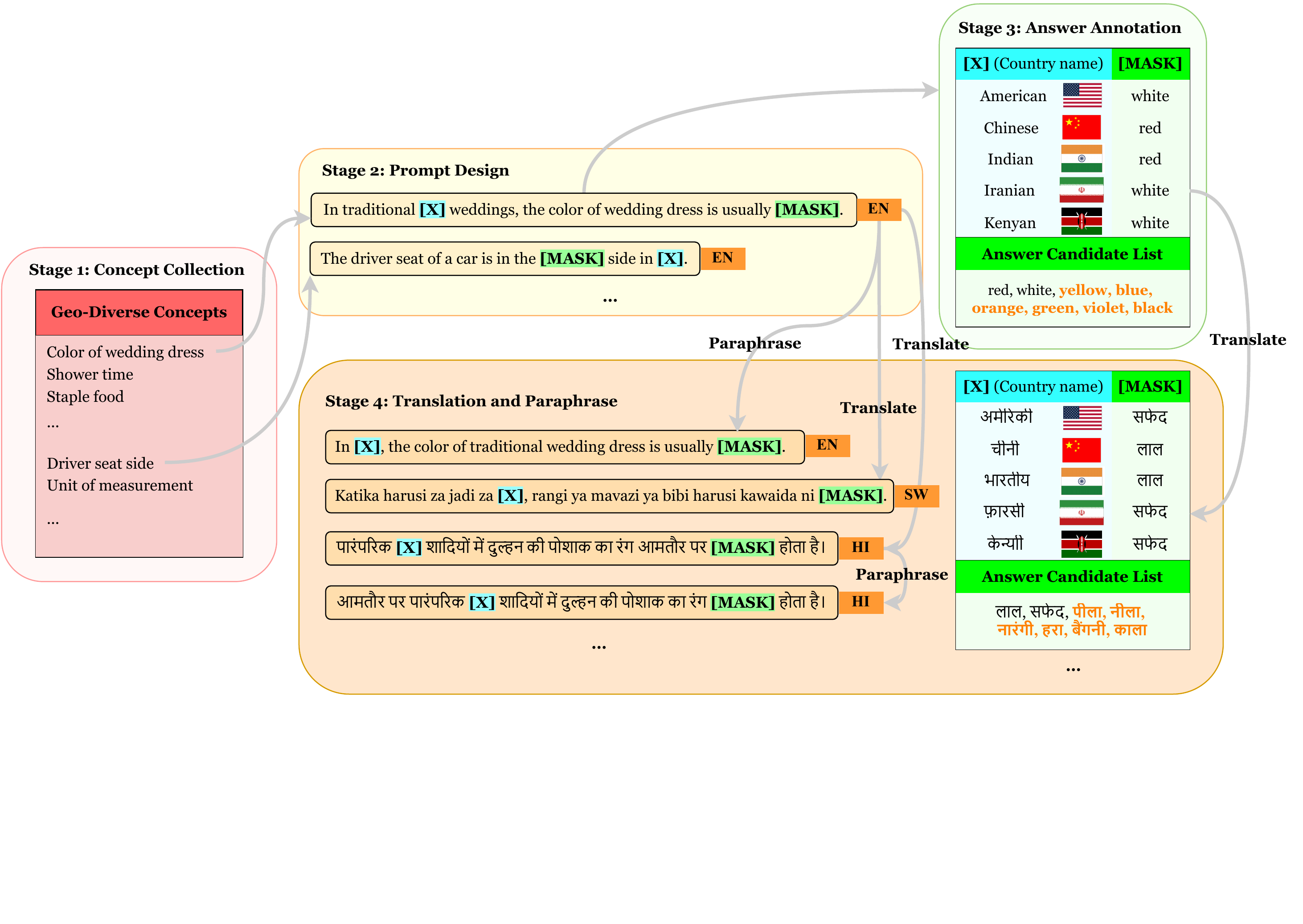} 
\caption{Overall annotation pipeline. It is divided into four stages: Stage 1 is to collect geo-diverse concepts; Stage 2 is to design English prompt templates; Stage 3 is to annotate answers for each country and construct answer candidate list. Stage 4 is to translate the English prompts and paraphrase the translated multilingual prompts. Here we showcase English and Hindi answer annotations for demonstration.}
\label{annotation}
\end{figure*}

\section{\textsc{GeoMLama} Benchmark Construction}
To build a geo-diverse commonsense probing benchmark, we recruit annotators from five different countries, the United States, China, India, Iran, and Kenya to participate in annotation. The annotation process is separated into four stages. 1) We first ask the annotators to list geo-diverse concepts. 2) Based on the collected concepts, we then require annotators to design masked geo-diverse prompt templates in English. 3) After specifying prompts with country names, we request annotators to provide correct answers and form answer candidate list for each prompt. 4) We translate the English prompts into other languages and paraphrase them. The overview of the annotation pipeline is illustrated in Figure~\ref{annotation}. 

\subsection{Geo-Diverse Concept Collection}
Geo-diverse concepts are the foundation of designing geo-diverse prompts. The criteria of selecting geo-diverse concepts are shown as follows:

\paragraph{Universality and Diversity across Cultures.}
We require that the scenarios regarding the collected concepts to be universal but diverse across the different cultures. ``\emph{Color of wedding dress}'' qualifies our criteria as \textit{wedding dress} is a universally understood entity where its color is diverse across different cultures.

\paragraph{Avoiding Concepts involving Region-Specific Terms.}
We avoid probing models about region-specific factual knowledge, e.g., festival names and president names of the countries, as these concepts usually involve uncommonly used tokens in certain languages and thus introduce another layer of complexity to make inference.

Finally, we consider topics that cover habits and personal choices, cultures and customs, policies and regulations, and geography for subsequent annotations. Details are shown in Appendix~\ref{concept-list}.

\subsection{Geo-Diverse Prompt Template Design}
Centered on the collected geo-diverse concepts, annotators design English version of geo-diverse prompt templates that will be later paraphrased and translated into multilingual prompts. Given one geo-diverse concept, e.g., ``\emph{color of wedding dress}'', the corresponding prompt template would be a masked sentence that inquires the missing color information, e.g., ``\emph{The color of wedding dress is usually} \colorbox[rgb]{ 0,  1,  0}{[MASK]}.'' Since we intend to probe knowledge about different countries using these prompts, we further insert phrases such as ``\emph{In} \colorbox[rgb]{ .2,  1,  1}{[X]}, '', ``\emph{In traditional} \colorbox[rgb]{ .2,  1,  1}{[X]} \emph{wedding,} '' to indicate the country knowledge to be probed. Here \colorbox[rgb]{ .2,  1,  1}{[X]} is either one of the country names (the United States, China, India, Iran, and Kenya), or one of the corresponding modifiers ( American, Chinese, Indian, Iranian, and Kenyan).

\subsection{Answer and Answer Candidate List Annotation}
For each masked geo-diverse prompt with a specified country name, we request the annotators to provide correct answers for the masked words. For instance, given a prompt about bridal outfit color in traditional Chinese weddings, ``\textit{In traditional Chinese weddings, the color of wedding dress is usually} [MASK]'', annotators are required to provide the answer ``\emph{red}'' for [MASK]. The answers are all provided by annotators who are familiar with the culture in one of our studied countries. Note that besides prompts with only one answer, some other prompts in \textsc{GeoMLama}, such as ``\textit{The staple food in Iran is} [MASK]'', can have \emph{multiple} correct answers (``\textit{rice}'' and ``\textit{bread}'') for a single prompt. To further validate the correctness of answers, we distributed a survey to collect responses for knowledge about respondents’ own countries. We collected 33 responses from the five countries, and retained the answers with majority support.

In this work, we focus on investigating whether PLMs are capable of predicting correct answers among all the possibilities of different countries. For example, we wonder if PLMs can predict the  dress color at Chinese wedding is ``\emph{red}'' over the other possibility, such as ``\emph{white}''. Therefore, we pair each prompt with an additional answer candidate list composed by the probable choices and multilingual PLMs are constrained to make predictions from the list. Specifically, each list contains the union of all correct answers of five countries and additional confounding candidates sharing the same word types with those correct answers. For the prompts about color of wedding dress, the union of correct answers is \{``\emph{red}'', ``\emph{white}''\}. Other than the two colors, as illustrated in Figure~\ref{annotation}, we also append confounders such as, ``\emph{yellow}'', ``\emph{black}'', ``\emph{blue}'' to the list (the orange letters in grids titled with ``Answer Candidate List''). The final answer candidate list for prompts about color of wedding dress will be \{``\emph{red}'', ``\emph{white}'', ``\emph{yellow}'', ``\emph{black}'', ``\emph{blue}'', ...\}. 
Note that the contents and lengths of answer candidate lists for prompts about different concepts vary greatly.

\subsection{Prompt Translation and Paraphrase}
We then obtain multilingual geo-diverse prompts via translating the annotated English prompts into four other languages Chinese, Hindi, Persian, and Swahili. We leverage Google Translation API to translate English prompts and each translated prompt is manually checked and corrected by annotators familiar with both English and any of the four studied languages. Besides, since it is shown that probing results are sensitive to small perturbation to the prompts~\cite{jiang-etal-2020-know}, we further generate four paraphrases for each prompt to obtain more robust probing results. Specifically, we paraphrase English prompts via a round of backtranslation\footnote{Based on Google Translation API.} in which we first translate English prompts to German ones and then translate them back to English. For prompts in other languages, their paraphrases are generated by backtranslation that translates texts to English and translate them back to the original languages. The paraphrases in a particular language are validated and modified by native speakers.

In total, we annotate 3125 prompts with answers and corresponding candidates in \textsc{GeoMLama}. All the prompts are designed based on 16 geo-diverse concepts listed in Appendix~\ref{concept-list}, and there are 625 prompts for each of the five languages. More details are described in Appendix~\ref{appendix_stats}.

\section{Probing Methods on \textsc{GeoMLama}}
\label{probing}
\citet{petroni-etal-2019-language} introduce the LAnaguage Model Analysis (LAMA) setup to probe knowledge stored in the pre-trained language models using masked templates. Without any additional fine-tuning, given a masked prompt, models are required to recover masked tokens with entities with the highest probability for the prompt context. Following LAMA probe, on \textsc{GeoMLama}, we study whether models are capable of seeking the most appropriate answers to from answer candidate list according to given geo-diverse prompts.

\citet{kassner-etal-2021-multilingual} follow LAMA probe to investigate entity knowledge in multilingual BERT only. In this work, we probe a diverse set of language models on \emph{geo-diverse commonsense knowledge} by scoring answer candidates and calibrating the score of each candidate.

\subsection{Scoring Answer Candidates}
We score answer candidates based on log likelihood of generating answer candidates given prompts. Different model families have their individual inference methods to obtain the scores. In the following, we introduce the probing methods for masked language models. Details of other probing methods on autoregressive and encoder-decoder language models are shown in Appendix~\ref{eval-main}.
\label{p_entity}

\paragraph{Masked Language Models (mBERT, XLM, XLM-R family).}
Given an answer candidate $e$ (e.g., ``\textit{chopsticks}'') that is tokenized into subtokens $e_1, e_2, ..., e_L$ (e.g., ``\textit{chop}'', ``\textit{stic}'', ``\textit{ks}'') such that $e_i \in V$ where $V$ is the vocabulary and $t$ is the prompt (e.g., ``\textit{In China}, \textit{people usually eat food with} [$\mathrm{MASK}_1$]...[$\mathrm{MASK}_L$].''), we assign a score $l_e$ based on the log probability of recovering the answer candidate $e$ in the masked prompt. Formally, $l_e$ is defined as
\begin{equation}
\label{eq:prob}
\small
 \frac{1}{L}\sum_{i=1}^{L} \log( p([\mathrm{MASK}_i]=e_i|[\mathrm{MASK}_{<i}]=e_{<i}, t)). \tag{1}
\end{equation}

According to Eq.\eqref{eq:prob}, we perform $L$ forward passes, each of which helps in obtaining conditional probability of generating one subtoken. 
To illustrate, $i^{th}$ forward pass inference would be $p([\mathrm{MASK}_i] = e_i | \text{ ``\textit{In China, people usually eat food with} } e_1 \text{ } e_2 \text{ }\\ ...e_{i-1} \text{ } [\mathrm{MASK}_i]...[\mathrm{MASK}_L]\text{''}).$

Here we further normalize the sum of log likelihood by the number of subtokens $L$ to help in reducing the effect of length. The other model families discussed in Appendix~\ref{eval-main} also adopt the normalization strategy.



\subsection{Calibrating Answer Candidates}
The way to score answer candidates $e \in \mathcal{E}$ (e.g., ``\textit{chopsticks}'' $\in$ \{``\textit{chopsticks}'', ``\textit{hands}'', ``\textit{spoons}'', ``\textit{knives}''\}) given the prompt $t$ for a country $C$ (e.g., ``\emph{In China, people usually eat food with} [MASK].'') is illustrated in \S \ref{p_entity}. However, this scoring mechanism is likely to be biased towards statistical correlations learned during pre-training~\cite{zhao2021calibrate} whilst ignoring the country-specific information present in the prompt. For instance, the model might choose ``\textit{knives}'' over 
``\textit{chopsticks}'' because ``\textit{knives}'' may occur more often than ``\textit{chopsticks}'' in pre-training corpora. Hence, we calibrate models with the prior probability of answer predictions in the absence of any country information. The final score given to each answer in the answers candidate set is given by:
\begin{align*}\label{eq:4}
    s_e = l_e-l'_e, \tag{2}
\end{align*}
where $l'_e$ is obtained using the same approach as $l_e$ but the input prompt for calculating $l'_e$ is the one without country information (e.g., ``\textit{People usually eat food with} [MASK].'' without ``\textit{In China,}'').

\begin{figure*}[htbp]
\centering
\begin{subfigure}{0.62\columnwidth}
\centering
\includegraphics[width=\textwidth]{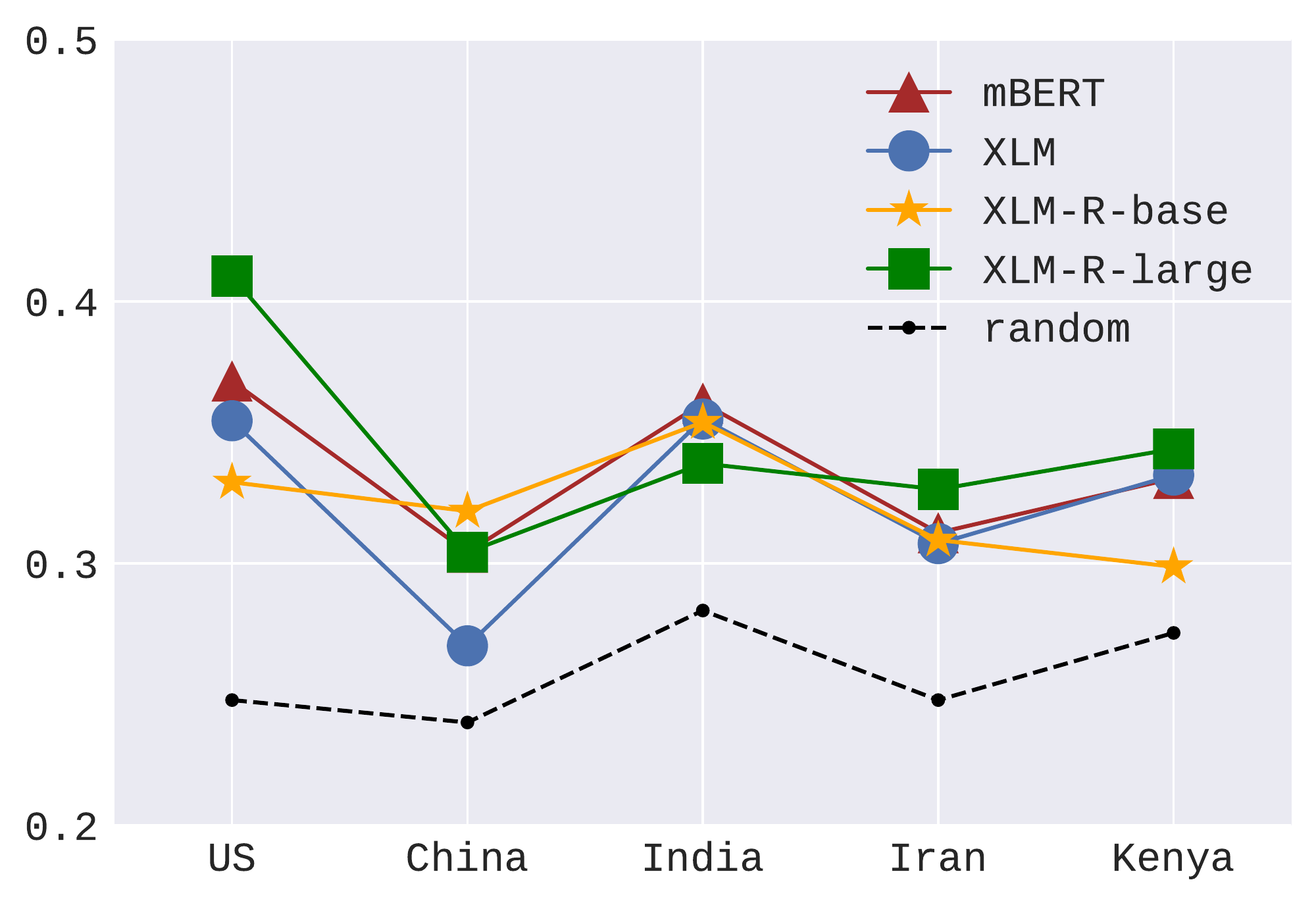}
\caption{mBERT, XLM, XLM-R family.}
\end{subfigure}
\hspace{12pt}
\begin{subfigure}{0.62\columnwidth}
\centering
\includegraphics[width=\textwidth]{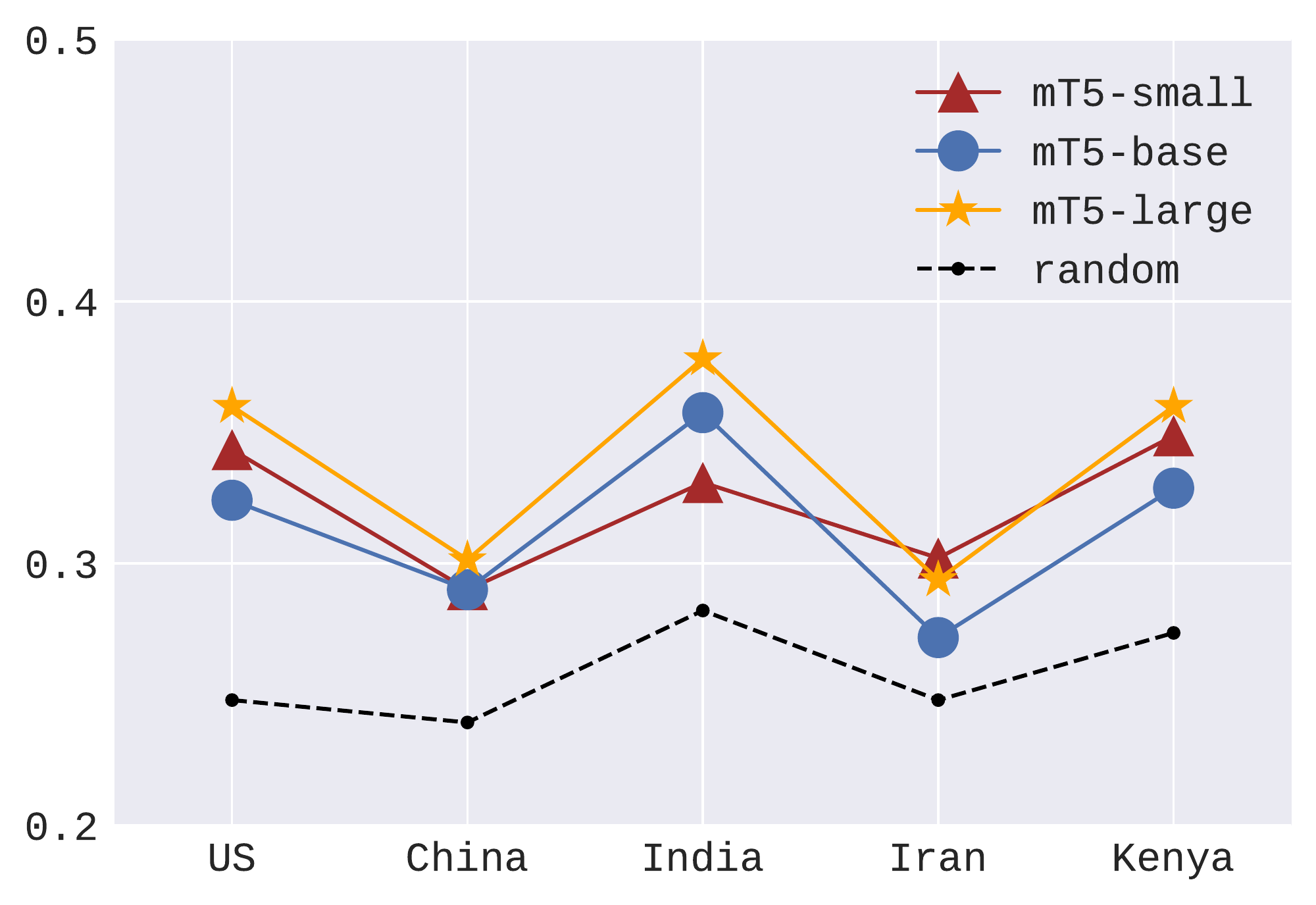}
\caption{mT5 family.}
\end{subfigure}
\hspace{12pt}
\begin{subfigure}{0.62\columnwidth}
\centering
\includegraphics[width=\textwidth]{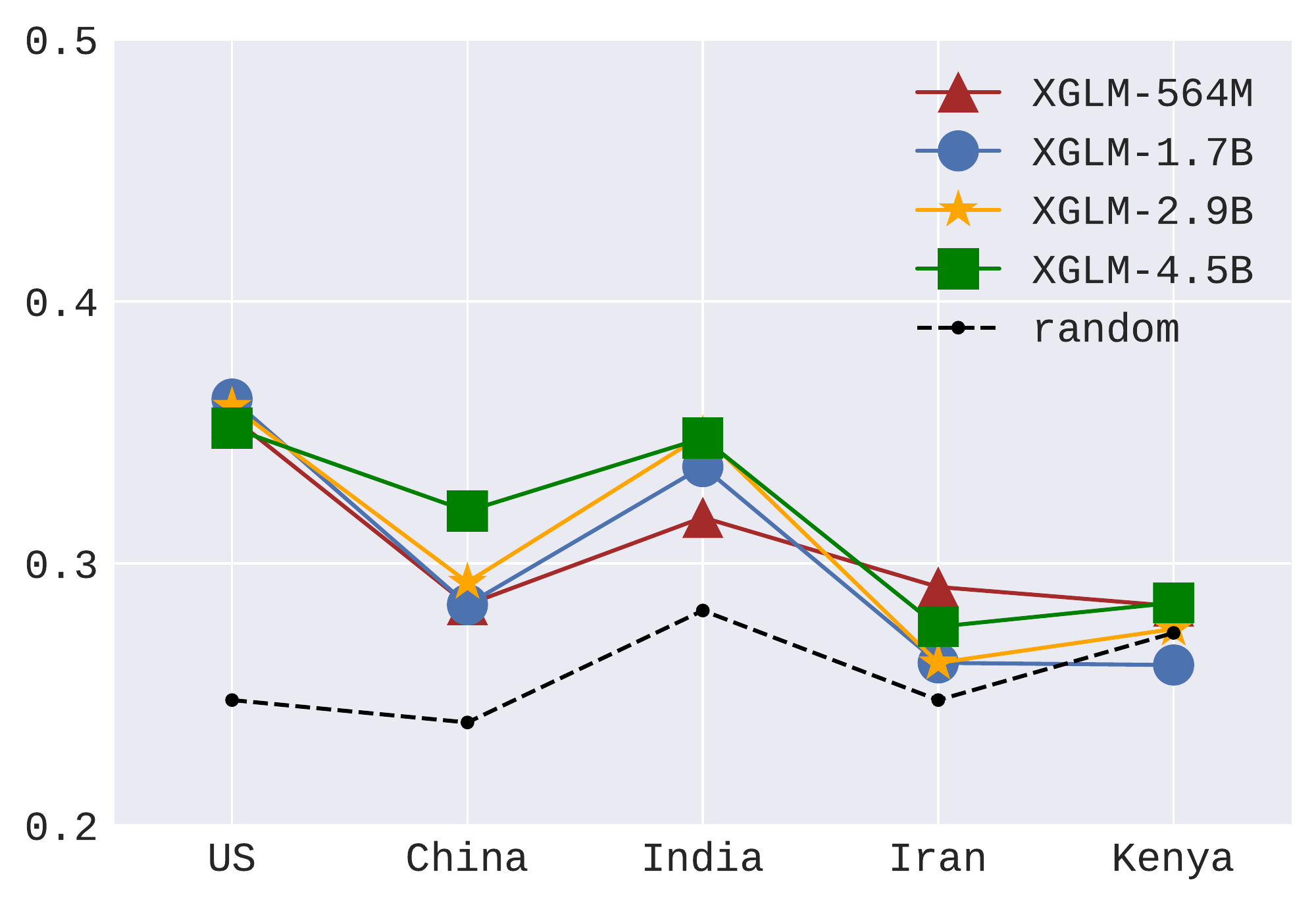}
\caption{XGLM family.}
\end{subfigure}
\caption{Multilingual PLMs' performance on probing knowledge about the studied countries averaged over all languages. Complete results are shown in Appendix~\ref{results-w}.}
\label{size-summ}
\end{figure*}

\begin{figure*}[htbp]
\centering
\begin{subfigure}{0.62\columnwidth}
\centering
\includegraphics[width=\textwidth]{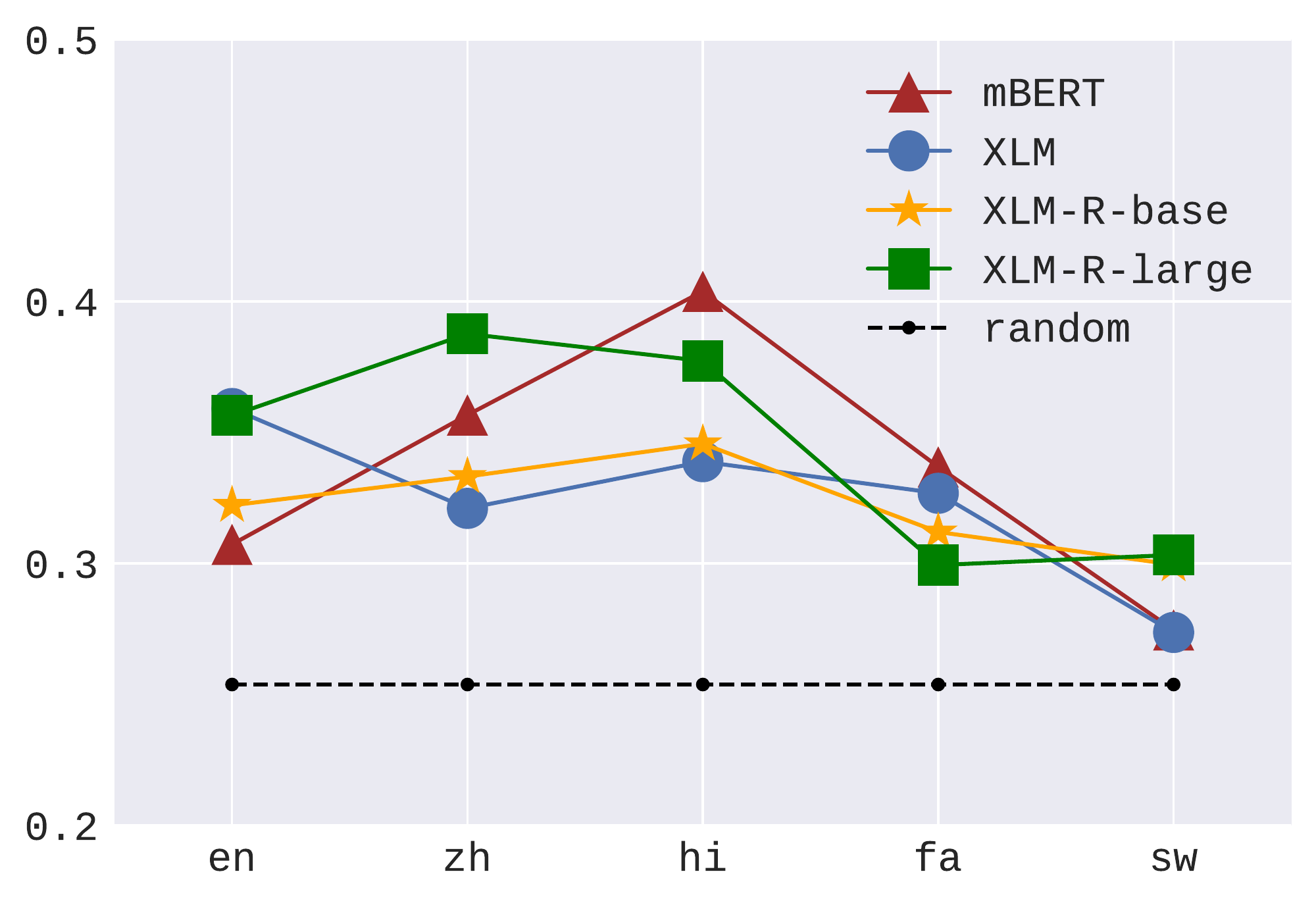}
\caption{mBERT, XLM, XLM-R family.}
\end{subfigure}
\hspace{12pt}
\begin{subfigure}{0.62\columnwidth}
\centering
\includegraphics[width=\textwidth]{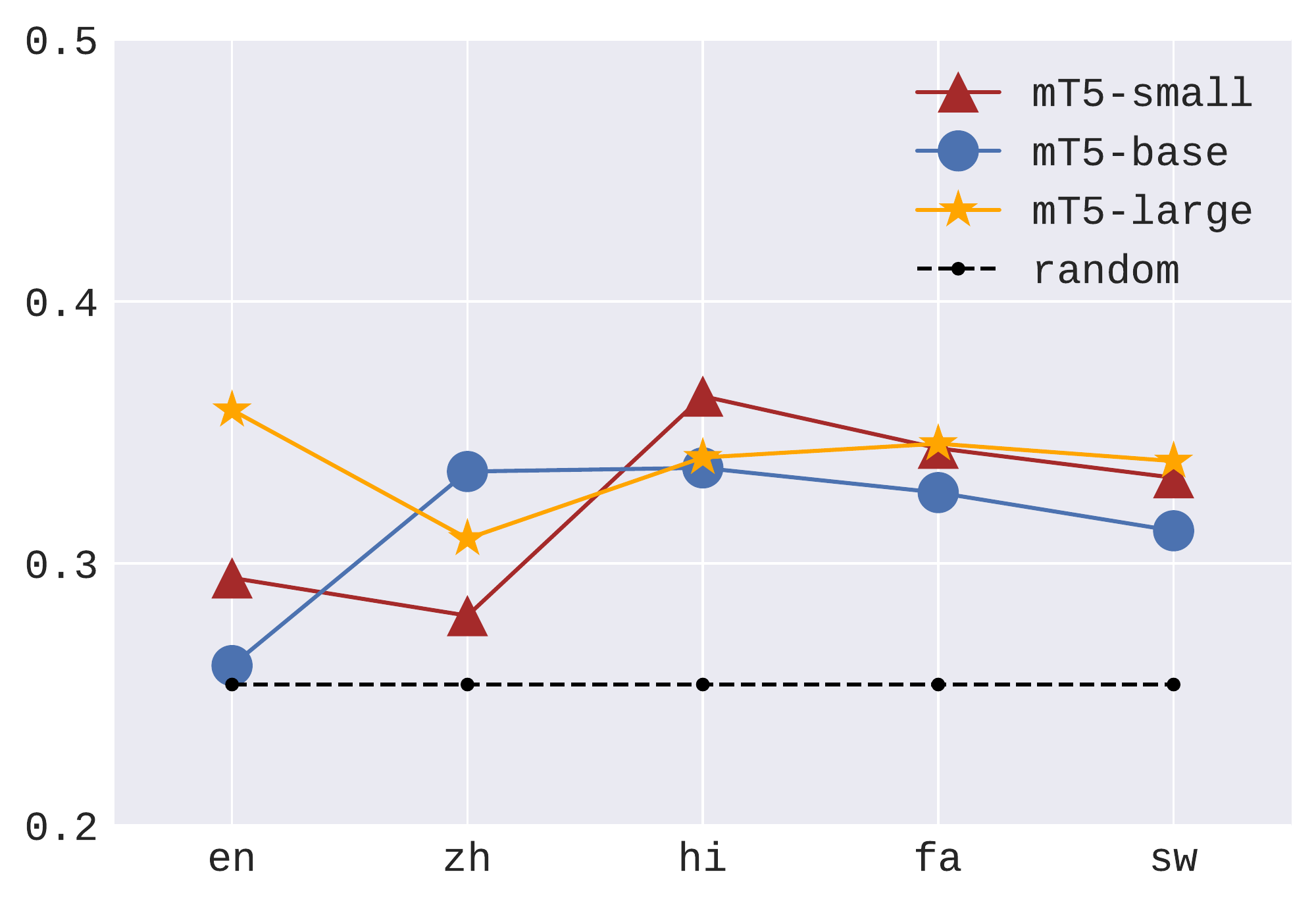}
\caption{mT5 family.}
\end{subfigure}
\hspace{12pt}
\begin{subfigure}{0.62\columnwidth}
\centering
\includegraphics[width=\textwidth]{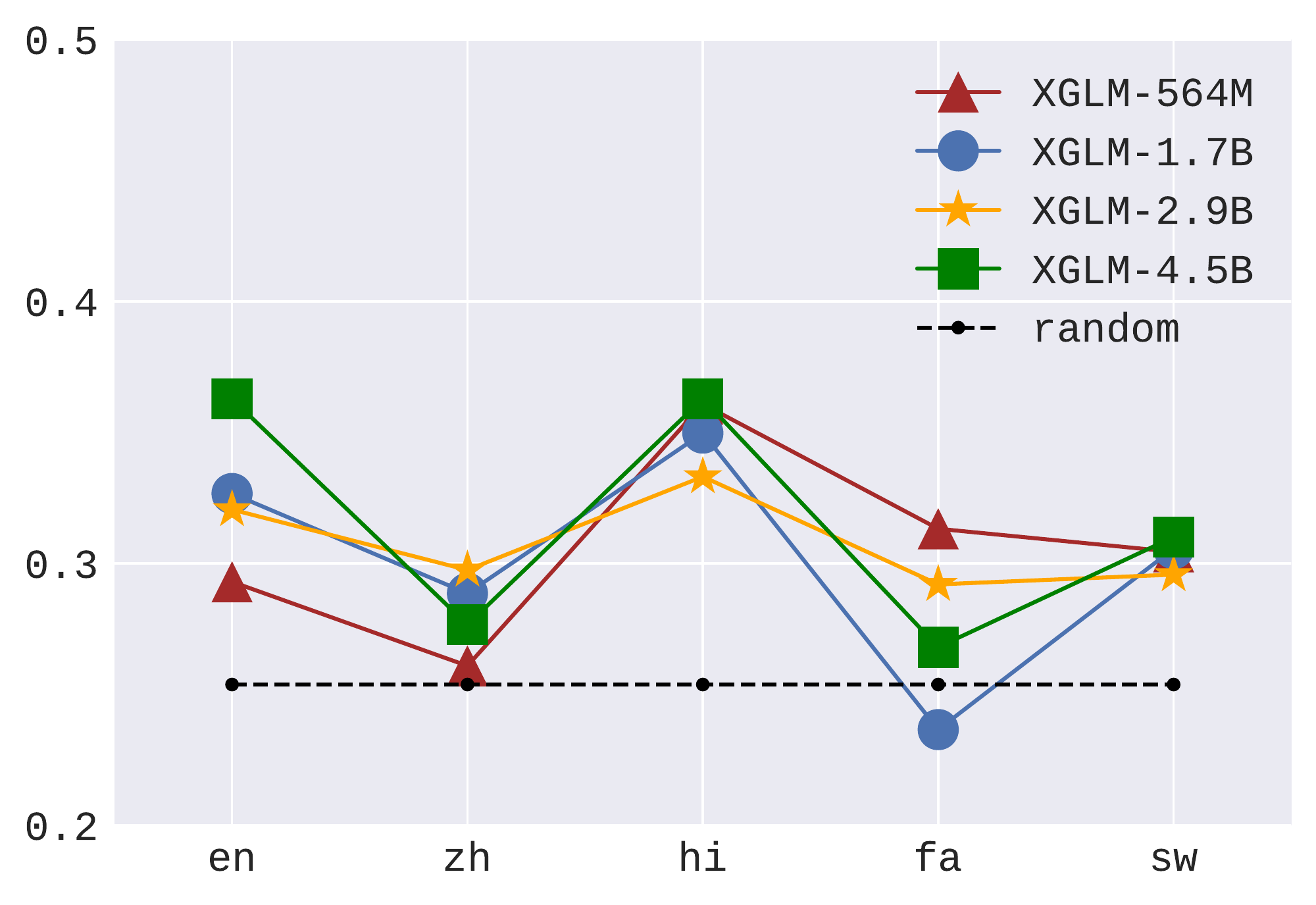}
\caption{XGLM family.}
\label{4c}
\end{subfigure}
\caption{Multilingual PLMs' performance averaged over countries when using multilingual prompts. ``\textit{en}'', ``\textit{zh}'', ``\textit{hi}'', ``\textit{fa}'', and ``\textit{sw}'' denote English, Chinese, Hindi, Persian, and Swahili. Complete results are shown in Appendix~\ref{results-w}.}
\label{lang-summ}
\end{figure*}

\subsection{Evaluation Metric}
\label{metric}
We use the ratio of total number of model's correct predictions to the total number of gold answers as model performance on \textsc{GeoMLama}. Specifically, given a prompt $t_i$ with $g_i$ gold answers, we count the number of top-$g_i$ model predictions that also appear in the gold answer list as $c_i$, based on the final score in Eq.\ref{eq:4}. For example, since there are two gold answers for the prompt ``\textit{The staple food in Iran is} [MASK]'', ``\textit{rice}'' and ``\textit{bread}'', $g_i=2$. In total, there are eight candidates in the answer candidate list \{``\emph{bread}'', ``\emph{noodles}'', ``\emph{rice}'', ``\emph{meat}'', ``\emph{maize}'', ...\} for this prompt. Assume one multilingual PLM assigns the highest $g_i$ scores to the candidates ``\emph{noodles}'' and ``\emph{rice}''. Then $c_i=1$, since only one of ``\emph{noodles}'' and ``\emph{rice}'' is the gold answer of the prompt. We then sum up all $c_i$ and $g_i$ to calculate the ratio, ${\sum^n_{i=1}c_i}/{\sum^n_{i=1}g_i},$ where $n$ is the total number of prompts in \textsc{GeoMLama}.

\section{Analysis on Multilingual PLMs}
In this section, we are interested in analyzing following questions: 1) Are bigger multilingual PLMs more geo-diverse than smaller ones? 2) In the absence of any particular country information in the prompts, are multilingual PLMs biased towards the knowledge towards certain countries? 
3) Can native language probe the knowledge about a particular country best? 
4) Given a particular language, can the corresponding country's knowledge be most accurately probed by the language?

To this end, we experiment with 11 multilingual PLMs\footnote{We also experiment with GPT-3 as it is also pre-trained on multilingual corpora. However, the results are not included in main paper because GPT-3 probing convention does not adopt cloze statements as the other 11 multilingual PLMs do. More setup details and results can be found in Appendix~\ref{eval-gpt3}.} including mBERT~\cite{devlin-etal-2019-bert}, XLM~\cite{conneau2019cross}, XLM-R family\footnote{XLM-R-base, XLM-R-large.}~\cite{conneau-etal-2020-unsupervised}, mT5 family\footnote{mT5-small, mT5-base, mT5-large.}~\cite{xue-etal-2021-mt5}, and XGLM family\footnote{XGLM-564M, XGLM-1.7B, XGLM-2.9B, XGLM-4.5B.}~\cite{lin2021few}.
We freeze pre-trained model parameters provided by HuggingFace Transformers~\cite{wolf-etal-2020-transformers} and do not fine-tune the models during probing.

\subsection{Overview of Model Performance}
Results are shown in Figure~\ref{size-summ} and~\ref{lang-summ}. Figure~\ref{size-summ} focuses on the comparison among performance of probing the knowledge about a particular country while Figure~\ref{lang-summ} compares the performance of using prompts in different languages. 

In Figure~\ref{size-summ}, we find that the performance of nearly all the multilingual PLMs lies in the range of 30\% to 40\% on probing each country's knowledge. Further, these multilingual PLMs significantly outperform random guess 2-15\%. It implies that multilingual PLMs can store geo-diverse commonsense knowledge and some stored knowledge can be accurately elicited even if we merely change the country names in the prompt. 

As illustrated in Figure~\ref{lang-summ}, we observe that the performance of using prompts in different languages is generally from 30\% to 40\% and higher than random guess 2-15\% as well. Moreover, we find that English and Hindi prompts are the most effective ones to probe geo-diverse knowledge, while Persian and Swahili prompts cannot achieve comparable results. In particular, from Figure~\ref{4c}, using Persian prompts to probe XGLM-1.7B leads to worse performance than random guess.

\subsection{Effect of Model Size}
According to \citet{petroni-etal-2019-language,roberts-etal-2020-much}, bigger models can generally store more knowledge and achieve better performance on downstream NLP tasks such as open-domain QA~\cite{joshi-etal-2017-triviaqa,kwiatkowski-etal-2019-natural}. To this end, we investigate whether larger models indeed perform better than the smaller ones on \textsc{GeomLAMA}. For a fair comparison, we only compare models in the same model families. 
This avoids comparing models with different pre-training corpora and learning objectives.

The comparison results over the three model families are shown in Figure~\ref{size-summ} and~\ref{lang-summ}. We observe that the larger models only perform marginally better than their smaller counterparts on \textsc{GeoMLAMA}. For the three model families, XLM-R, mT5, and XGLM, the performance gap between the largest and smallest models on all the prompts in \textsc{GeoMLama} is merely 2.23\%, 2.42\%, and 1.46\%, respectively. In specific cases (e.g., probing XGLM family using Persian prompts), the largest model can be even worse than its smallest variant. It demonstrates that even if large models have nearly an order of magnitude more parameters than small models, large models cannot store geo-diverse commonsense significantly better than small models. This highlights that \textsc{GeoMLama} is a challenging task and being better on the standard multilingual NLP tasks does not guarantee good performance.

\subsection{Intrinsic Model Bias without Country Information}
Each prompt in \textsc{GeoMLama} consists of the country information. 
However, it is still not clear as to what information is probed innately when we query multilingual PLMs without any country information. To study this phenomenon, we further probe multilingual PLMs with the prompts where the country token is removed. For example, instead of ``\textit{In traditional} \colorbox[rgb]{ .2,  1,  1}{Kenyan} \textit{weddings, the color of wedding dress is usually} [MASK]'', we implement a new round of probing with the pruned prompt, ``\textit{In traditional weddings, the color of wedding dress is usually} [MASK]''. The new prompts can elicit the knowledge that multilingual PLMs are intrinsically inclined towards predicting.

\begin{figure*}[htbp]
\centering
\begin{subfigure}{0.62\columnwidth}
\centering
\includegraphics[width=\textwidth]{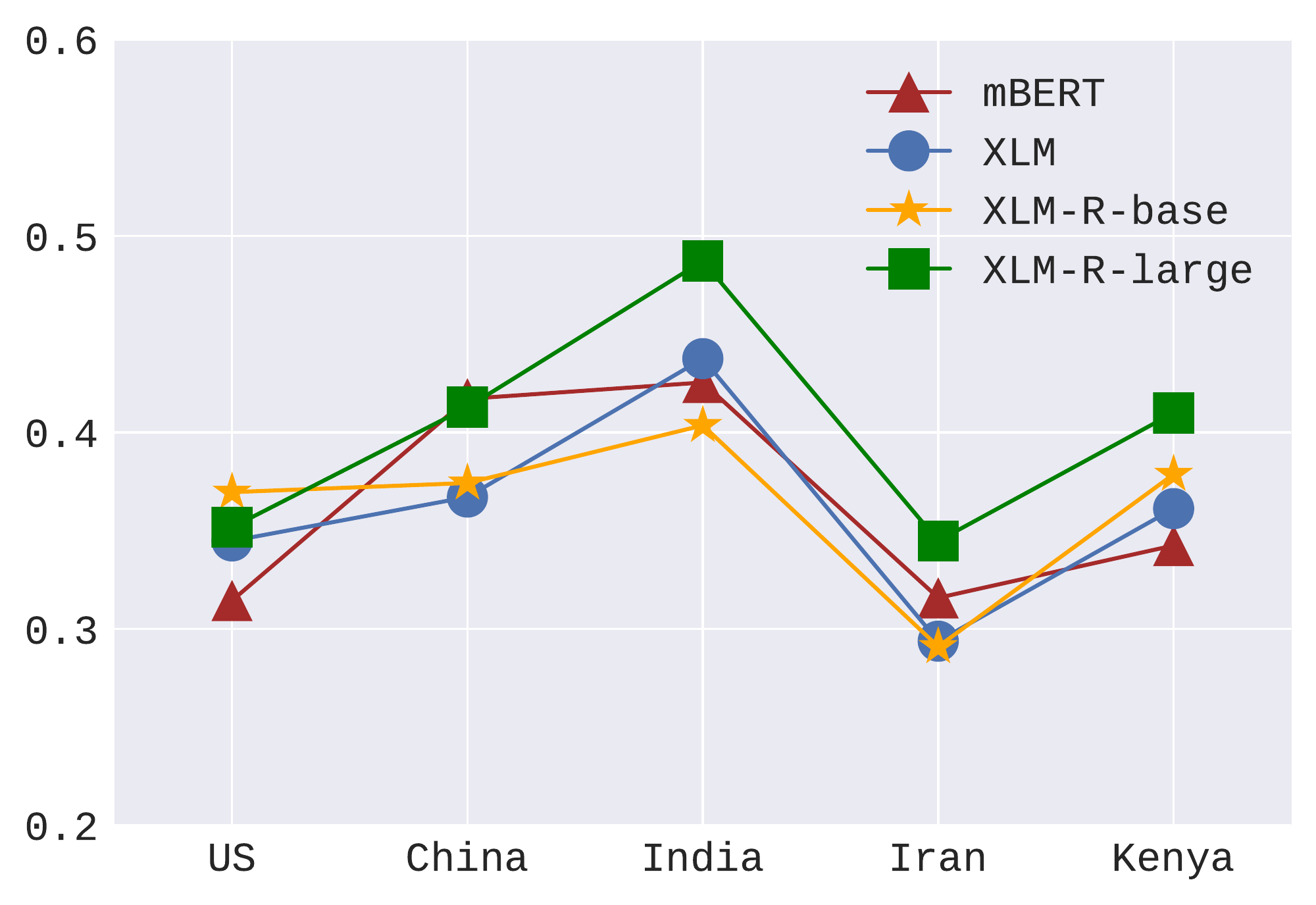}
\caption{mBERT, XLM, XLM-R family.}
\end{subfigure}
\hspace{12pt}
\begin{subfigure}{0.62\columnwidth}
\centering
\includegraphics[width=\textwidth]{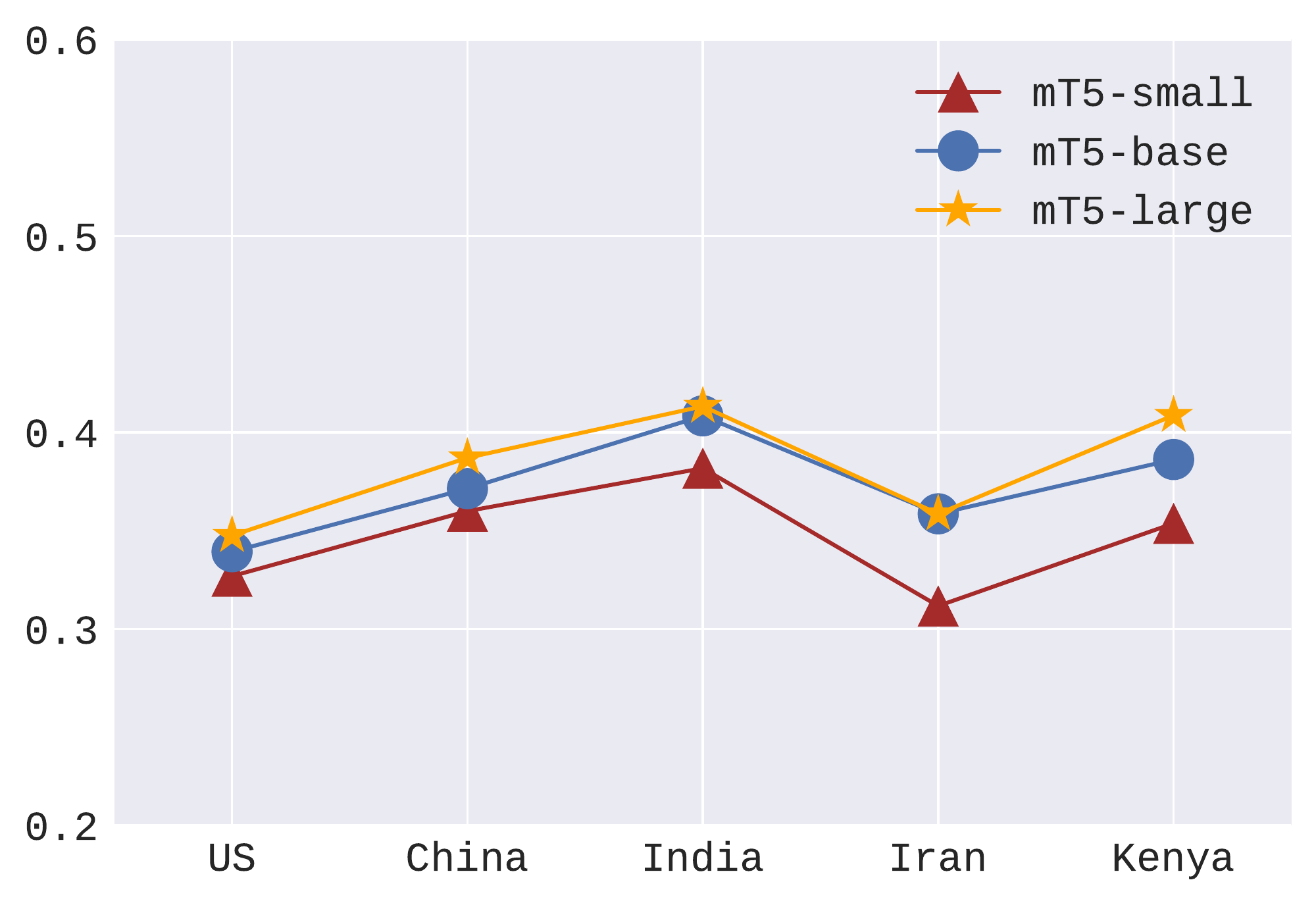}
\caption{mT5 family.}
\end{subfigure}
\hspace{12pt}
\begin{subfigure}{0.62\columnwidth}
\centering
\includegraphics[width=\textwidth]{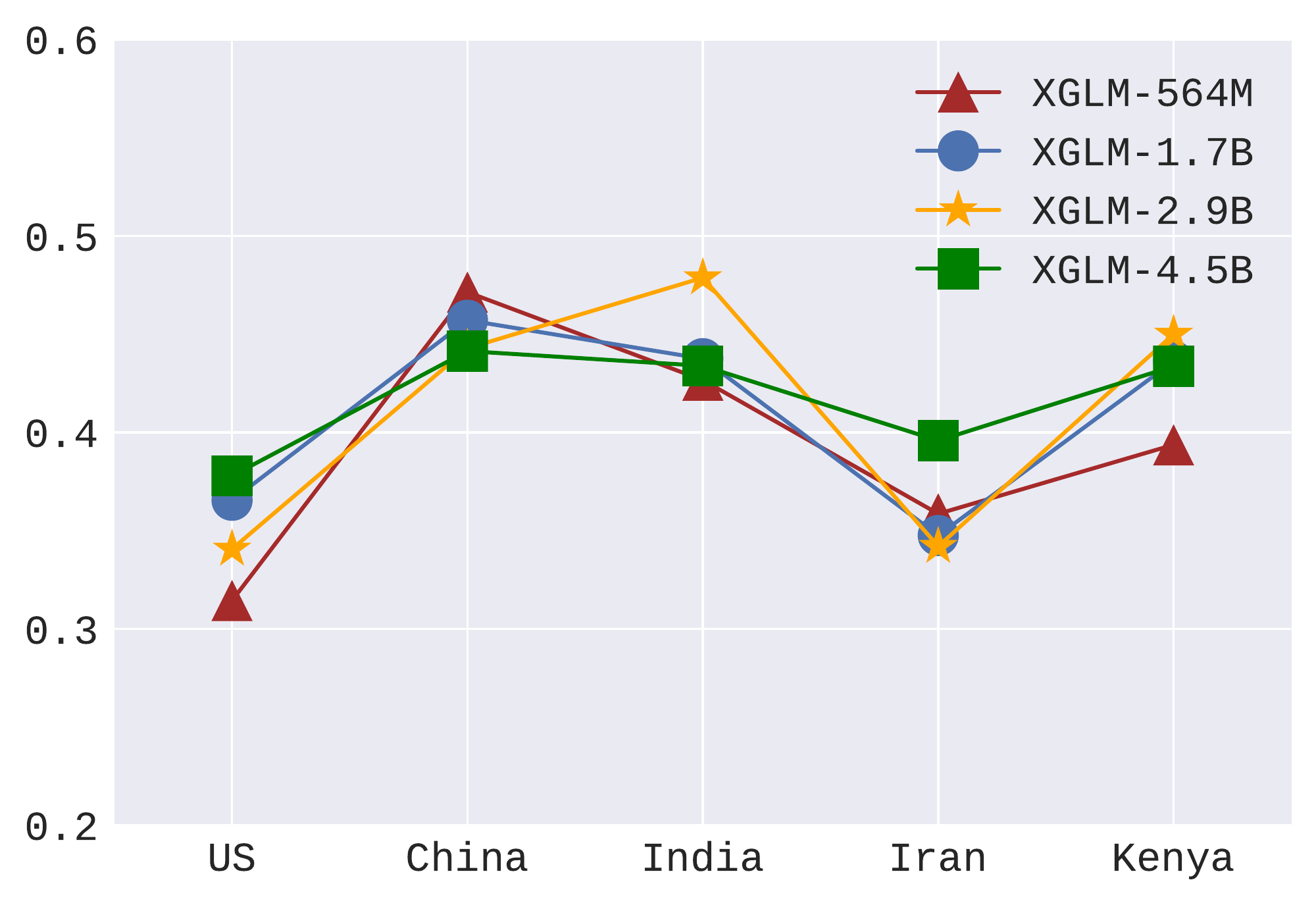}
\caption{XGLM family.}
\end{subfigure}
\caption{Average performance of multilingual PLMs when fed with prompts without any specified country names. Complete results are shown in Appendix~\ref{results-wo}.}
\label{fig:bias-summ}
\end{figure*}

\begin{table}[]
\centering
\scalebox{0.72}{
\begin{tabular}{lccccc}
\toprule
\textbf{Models}         & \textbf{US} & \textbf{China} & \textbf{India} & \textbf{Iran} & \textbf{Kenya} \\
\midrule
\textbf{mBERT}          & fa          & sw             & en             & fa            & zh             \\
\midrule
\textbf{XLM}            & fa          & en             & en             & zh            & zh             \\
\midrule
\textbf{XLM-R-base}     & fa          & zh             & zh             & fa/sw         & en             \\
\textbf{XLM-R-large}    & fa          & zh             & en             & en            & zh             \\
\midrule
\textbf{mT5-small}      & fa          & en             & en             & sw            & sw             \\
\textbf{mT5-base}       & fa          & en             & zh             & hi            & sw             \\
\textbf{mT5-large}      & fa          & sw             & sw             & fa            & hi             \\
\midrule
\textbf{XGLM-564M}      & fa          & en             & sw             & fa            & fa/hi          \\
\textbf{XGLM-1.7B}      & fa          & sw             & en             & fa            & fa             \\
\textbf{XGLM-2.9B}      & fa          & en             & en             & hi            & fa             \\
\textbf{XGLM-4.5B}      & fa          & zh             & en             & fa            & en             \\
\midrule
\textbf{Best Languages} & \textbf{fa} & \textbf{en}    & \textbf{en}    & \textbf{fa}   & \textbf{zh/fa} \\
\bottomrule
\end{tabular}
}
\caption{Best languages to probe each country's knowledge. Each language in the last row ``\textbf{Best Languages}'' is the one appearing most in its located column.}
\label{bestlang}
\end{table}

\begin{table}[]
\centering
\scalebox{0.75}{
\begin{tabular}{lccc}
\toprule
\textbf{Words}              & \textbf{English}  & \textbf{Chinese} & \textbf{Swahili} \\
\midrule
\textbf{rice, staple food}      & 1040  & 33 & 7                   \\
\midrule
\textbf{bread, staple food} & 33 & 37 & 1 \\
\bottomrule
\end{tabular}
}
\caption{Word co-occurrence of ``\textit{rice}'', ``\textit{bread}'' and ``\textit{staple food}'' in English, Chinese and Swahili Wikipedia, respectively.
}
\label{tab:wocountrycase}
\end{table}

As shown in Figure~\ref{fig:bias-summ}, we find that for most multilingual PLMs, the knowledge about India is captured frequently in the absence of any country information. Whereas, knowledge about the United States is not well probed. It shows that at least, multilingual PLMs are not originally biased towards knowledge about Western countries like US.

We do a quantitative case study to further explain the phenomenon. We take a geo-diverse concept ``\textit{staple food}'' as an example. Rice and bread are the staple foods in China and the United States, respectively. According to Table~\ref{tab:wocountrycase}, in English, Chinese and Swahili Wikipedia, we find that the co-occurrence of ``\textit{staple food}'' and ``\textit{rice}'' is comparable or even way higher than ``\textit{staple food}'' and ``\textit{bread}''. It demonstrates that the popularity of Western knowledge across the world does not necessarily mean higher frequency in knowledge sources like Wikipedia. This may lead the models to predicting non-Western knowledge more precisely.

\subsection{Best Languages to Probe Knowledge about Countries}

In \textsc{GeoMLama}, prompts in different languages are used to probe knowledge about different countries. It is imperative to ask whether we elicit most knowledge about a country if we query the PLM with its native language. From Table~\ref{bestlang}, contrary to our intuition, the native language is not the best language to query its knowledge for most of the countries. In particular, Iran is the only country for which its native language Persian can help in drawing out maximum knowledge about it. For the United States and Kenya, the best probing language is Persian and for China and India, the best language is English. 

\begin{table}[]
\centering
\scalebox{0.66}{
\begin{tabular}{lccccc}
\toprule
\textbf{Models}         & \textbf{en}    & \textbf{zh}    & \textbf{hi}    & \textbf{fa} & \textbf{sw}    \\
\midrule
\textbf{mBERT}          & India          & India          & US             & US          & China          \\
\midrule
\textbf{XLM}            & India          & Kenya          & India          & US          & Kenya          \\
\midrule
\textbf{XLM-R-base}     & India          & China          & India          & US          & India          \\
\textbf{XLM-R-large}    & India          & US             & US             & US          & Kenya          \\
\midrule
\textbf{mT5-small}      & India          & Kenya          & Kenya          & US          & Kenya          \\
\textbf{mT5-base}       & India          & India          & Kenya          & US          & Kenya          \\
\textbf{mT5-large}      & India          & Kenya          & Kenya          & US          & India          \\
\midrule
\textbf{XGLM-564M}      & China          & US             & India/Kenya    & US          & India          \\
\textbf{XGLM-1.7B}      & India          & India          & India          & US          & India          \\
\textbf{XGLM-2.9B}      & India          & India          & US             & US          & India          \\
\textbf{XGLM-4.5B}      & India          & US/China/India & India          & US          & India          \\
\midrule
\textbf{Best Countries} & \textbf{India} & \textbf{India} & \textbf{India} & \textbf{US} & \textbf{India} \\
\bottomrule
\end{tabular}
}
\caption{Countries best probed with prompts in different languages. Each country in the last row ``\textbf{Best Countries}'' is the one appearing most in its located column.}
\label{bestcountry}
\end{table}

We speculate that our observations might be attributed to the reporting bias phenomenon \cite{grice1975logic, gordon2013reporting}. It is categorized by people rarely stating the obvious knowledge that is shared by everyone (commonsense) explicitly in the text. For instance, the fact that all the \textit{humans can murder} is disproportionately over-reported than \textit{humans can breathe} in the English text. 
This unbalanced frequency would lead to bias towards acquiring uncommon event knowledge from PLMs, instead of commonsense knowledge~\cite{shwartz-choi-2020-neural}. 
In our setting, we believe that reporting bias is a key ingredient in explaining our observed trends. For instance, indigenous population is less likely to record obvious facts about their culture in their native language texts as compared to the facts from other cultures. For example, when mentioning the driver seat side in India, compared with people living in other countries, Indian people will not talk too much about this because it is too trivial for them.

\begin{table}[]
\centering
\scalebox{0.6}{
\begin{tabular}{lcc}
\toprule
\textbf{Words}              & \textbf{Freq. of Co-occur}                   & \textbf{\# Co-occur} \\
\midrule
\textbf{rice, staple food, China}      &  3.6x   & 25                     \\
\textbf{bread, staple food, US} & 1x  & 7 \\
\midrule
\textbf{\begin{CJK*}{UTF8}{gbsn}米饭 (rice), 主食 (staple food), 中国 (China)\end{CJK*}}        &  3.2x    & 3                     \\
\textbf{\begin{CJK*}{UTF8}{gbsn}面包 (bread), 主食 (staple food), 美国 (US)\end{CJK*}}     &   3.2x      & 3 \\
\bottomrule
\end{tabular}
}
\caption{Word co-occurrence and frequency in English and Chinese Wikipedia. 
English Wikipedia has 72484142 sentences, 7.6 times more than those of Chinese Wikipedia, 9502859 sentences. `$n$x' denotes the frequency rate is $n$ times higher than the lowest one.
}
\label{reportbias}
\end{table}

We seek a quantitative evidence in the context of \textit{staple food} as a concept again to support our claim. Throughout the English and Chinese Wikipedia corpora, we count the co-occurrence of words ``\textit{China}'', ``\textit{rice}'' and ``\textit{staple food}'', and ``\textit{the United States}'', ``\textit{bread}'' and ``\textit{staple food}'' in their respective languages. The counting results are shown in Table~\ref{reportbias}. We notice that when China is mentioned, English words ``\textit{rice}'' and ``\textit{staple food}'' co-occur 25 times whereas it is mentioned merely 3 times in Chinese Wikipedia. Furthermore, in the context of the US, English words ``\textit{bread}'' and ``\textit{staple food}'' appear 7 times simultaneously while Chinese words \begin{CJK*}{UTF8}{gbsn}``面包 (\textit{bread})'' and ``主食 (\textit{staple food})''\end{CJK*} co-occur 3 times. Although the number of co-occurrence is higher in the English Wikipedia, the frequency rate of the Chinese word co-occurrence is 3.2 times higher, since the Chinese Wikipedia corpus is 7.6 times smaller than the English corpus. 
In summary, it shows that commonsense knowledge about a country is not mentioned more frequently in its native language corpus but might have higher occurrences in some other languages.

\subsection{Countries Best Probed with Prompts in Different Languages}
Apart from the best languages to probe knowledge about countries, conversely, we can also study the countries best probed with prompts in different languages. Specifically, we focus on the following question: 
Given one studied language $X$, is the country best probed the same as the indigenous country of language $X$?

We present our results in Table~\ref{bestcountry}. 
We observe that except Hindi, the countries best probed are distinct to the corresponding countries of language. For example, Swahili prompts probe Indian knowledge best instead of Kenya, and Persian prompts probe US knowledge best instead of Iran. It is also counter-intuitive because it is natural for people to imagine that the best probed country should be the one where a particular language is spoken most commonly. 

We can also ascribe the phenomenon observed for Q2 to the reporting bias. 
To analyze this observation, we compare the occurrence of knowledge about different countries in the same language corpus. We find that English words ``\textit{bread}'', ``\textit{staple food}'' and ``\textit{the United States}'' co-occur much less frequently than ``\textit{rice}'', ``\textit{staple food}'' and ``\textit{China}''. Besides, Chinese words \begin{CJK*}{UTF8}{gbsn}``面包 (\textit{bread})'', ``主食 (\textit{staple food})'' and ``美国 (\textit{the United States})'' co-occur 3 times, which is the same as co-occurrence of ``米饭 (\textit{rice})'', ``主食 (\textit{staple food})'' and ``中国 (\textit{China})''\end{CJK*}. The comparison results indicate that given one language, local country's knowledge may not appear the most, compared with knowledge about other countries.

\section{Conclusions}
We propose a knowledge probing benchmark, \textsc{GeoMLama}, to evaluate the extent of multilingual PLMs to store geo-diverse commonsense. Results show that multilingual PLMs can achieve significantly higher performance than random guess, suggesting that they are capable of storing geo-diverse knowledge. We also find that fed with prompts without any country cues, multilingual PLMs are not intrinsically biased towards knowledge about the United States. We further investigate the best language to probe the knowledge about a particular country, and the country best probed with prompts in a certain language. Surprisingly, we notice that the best language is not the country's native language, and the best probed country is not the indigenous country of the language. We connect this to reporting bias issue in geo-diverse context: one country's commonsense is seldom recorded in the text by people living in that country as it is too trivial and not worth mentioning for them. 


\section*{Acknowledgement}
We thank annotators for tremendous efforts on annotation and evaluation. We also greatly appreciate Tao Meng, Xiao Liu, Ashima Suvarna, Ming Zhong, Kuan-Hao Huang, I-Hung Hsu and other members of UCLA-NLP group for their helpful comments. 
This work was partially supported by NSF IIS-1927554, Sloan Research Fellow, Amazon AWS credits, Amazon Fellow, and a
DARPA MCS program under Cooperative Agreement N66001-19-2-4032. The views and conclusions are those of the authors and should not reflect the official policy or position of DARPA or the U.S. Government.

\section*{Limitations}
\textsc{GeoMLama} is proposed for evaluating the degree of potential geographic bias in multilingual PLMs. However, due to the limited coverage of countries, languages and geo-diverse concepts, \textsc{GeoMLama} may introduce unwanted bias. In \textsc{GeoMLama}, we only consider five countries and their native languages, which merely occupy a tiny portion of all the countries in the world and thousands of languages. Also, in countries like India, there are multiple commonly used languages, we limit our study on Hindi and will extend to more languages to study the phenomenon. Besides, we design prompts simply based on 16 general geo-diverse concepts. The extension on existing \textsc{GeoMLama} can help in obtaining more solid results and mitigating bias against uncovered countries and languages.

In this work, we mainly focus on evaluating multilingual PLMs on \textsc{GeoMLama} without studying how multilingual pre-training process affects the model performance on geo-diverse commonsense probing. We intend to explore effect of the process on model's geo-diversity in future work. Specifically, we aim to examine whether pre-training on multilingual corpora really brings more geo-diversity than pre-training on monolingual corpora does. Besides, we do not cover how to improve model performance on \textsc{GeoMLama} and other related tasks. We expect to seek approaches to improving model's geo-diversity while maintaining multilingual PLMs' performance on various multilingual benchmarks in future work as well.

\section*{Ethical Consideration}
As we propose a new benchmark in this paper, we provide details about compensation rate for annotators. We recruit five countries' college students and annotators from Amazon MTurk. We provide a fair compensation rate with \$12 per hour and in total around \$150 to the annotators on both prompt design, translation and evaluation. Note that part of annotations are done by the authors of this work.

\bibliography{anthology,custom}

\clearpage

\appendix

\section*{Appendix}

\section{Geo-Diverse Concept List}
\label{concept-list}

The general geo-diverse concepts are shown in Table~\ref{tab:list}. We summarize all the concepts into 16 general ones, covering rules, policies, geography, customs, personal choices and habits. Multiple prompts can be designed for each geo-diverse concept. For example, measurement units can involve units measuring height, weight and temperature, and thus annotators can create multiple prompts about various types of measurement units.

\section{Statistics of \textsc{GeoMLama}}
\label{appendix_stats}
Table~\ref{stats} shows the statistics of \textsc{GeoMLama}. In total, there are 3125 prompts in \textsc{GeoMLama}, 625 prompts about each country's knowledge. We also manifest the average numbers of gold answers and corresponding answer candidates for prompts regarding each country. Overall, the number of gold answers is 1.20 per prompt, with answer candidate list of average length 4.76. Here note that for prompts under the same topic (e.g., ``\textit{In traditional} \colorbox[rgb]{ .2,  1,  1}{[X]} \textit{weddings, the color of wedding dress is usually} [MASK].''), regardless of the exact country filled in \colorbox[rgb]{ .2,  1,  1}{[X]}, the answer candidate lists are the same for all the five countries. Therefore, the average length of answer candidates is identical to all the studied countries.

\section{Details of Evaluation Methods on Autoregressive and Encoder-Decoder Language Models}
\label{eval-main}

\paragraph{Autoregressive Language Models (XGLM family).} For autoregressive language models such as XGLM, we first replace masked token in the prompt with answer candidate tokens (e.g., ``\textit{In China, people usually eat food with} [MASK].''->``\textit{In China, people usually eat food with chopsticks.}''). The joint probability of generating all the tokens in the complete sentence is used for scoring answer candidates. Given a prompt template $t$ filled with an answer candidate $e$, t is tokenized into $K$ tokens (e.g., $t_1$, $t_2$, ..., $t_K$). We assign score $l_e$ to the answer candidate as:
\begin{equation}\label{eq:2}
    l_e=\frac{1}{K}\sum_{i=1}^{i=K} \log (p(t_i | t_{<i})). \tag{3}
\end{equation}
Here, we perform $K$ forward passes to the autoregressive language model to obtain log probability of generating the whole sentence with the answer candidate $e$.
In this case, the $i^{th}$ forward pass inference would calculate $p(t_i | \text{ ``\textit{In China,} } ..., t_{i-2} \text{ } t_{i-1} \text{''}).$

\begin{table}[]
\centering
\scalebox{0.6}{
\begin{tabular}{lccc}
\toprule
\textbf{Countries} & \# \textbf{Prompts}           & \textbf{\# Avg. Gold Answers} & \textbf{\# Avg. Answer Candidates} \\ \midrule
\textbf{US}        & \multirow{5}{*}{625} & 1.16                & \multirow{5}{*}{4.76}     \\
\textbf{China}     &                      & 1.12                &                           \\
\textbf{India}     &                      & 1.32                &                           \\
\textbf{Iran}      &                      & 1.16                &                           \\
\textbf{Kenya}     &                      & 1.25                &                           \\ \midrule
\textbf{Overall}   & 3125                 & 1.20                & 4.76       \\
\bottomrule
\end{tabular}
}
\caption{Detailed statistics of \textsc{GeoMLama}.}
\label{stats}
\end{table}

\paragraph{Encoder-Decoder Language Models (mT5 family).} During pre-training of encoder-decoder language models mT5, a masked sequence is input to encoder, and decoder learns to recover the $L$ masked tokens in autoregressive fashion. Therefore, we input a masked prompt $t$ into the models (e.g., ``\textit{In China, people usually eat food with }$[\mathrm{MASK}]$'') and calculate the score for answer candidate $e$ as:
\begin{equation}\label{eq:3}
    l_e=\frac{1}{L}\sum_{i=1}^{i=L} \log (p(e_i | e_{<i}, t)). \tag{4}
\end{equation}
Computing Eq.\ref{eq:3} requires $L$ forward passes, since the decoder needs to generate $L$ tokens. Here $i^{th}$ forward pass inference would be $p(e_i | e_1 \text{ } e_2 ... e_{i-1}, \text{ ``\textit{In China, people usually eat}} \\ \text{\textit{food with }} [\mathrm{MASK}]\text{''}).$ Note that mT5 can use one single $[\mathrm{MASK}]$ token to represent multiple consecutive masked tokens. Thus, different from masked language models, mT5 models are simply fed with the prompt with only one $[\mathrm{MASK}]$ token instead of $L$ $[\mathrm{MASK}]$ tokens.

\section{Evaluating GPT-3 on \textsc{GeoMLama}}
\label{eval-gpt3}
Approach to probing GPT-3 is different from the methods mentioned in \S \ref{probing}. Instead of feeding declarative prompt sentences, we leverage Question Answering (QA) API empowered by GPT-3 and input questions to query the knowledge. For example, instead of using ``\textit{In traditional Chinese weddings, the color of wedding dress is usually} [MASK]'', we first convert it to question form like ``\textit{What is the color of wedding dress in an American wedding?}'' and query GPT-3 with the converted question. During evaluation stage, rather than scoring answers from given answer candidate list, GPT-3 can generate open-ended answers and we evaluate GPT-3 predictions using the same metric in \S \ref{metric}. Considering the huge time cost of manually inputting questions by annotators to GPT-3 API, we do not convert paraphrased prompts to questions and perform analysis on them. In other words, the number of tested questions is only $1/5$ out of the total number of prompts in \textsc{GeoMLama}, which is 625. 

\begin{table}[]
\centering
\scalebox{0.82}{
\begin{tabular}{lc}
\toprule
\textbf{Categories}                                          & \textbf{Concepts}    \\
\midrule
\multirow{5}{*}{\textbf{rules, policies, geography}}         & traffic rules        \\
                                                             & measurement units    \\
                                                             & date formats          \\
                                                             & color of stock price \\
                                                             & climate              \\
\midrule
\multirow{11}{*}{\textbf{customs, personal choices, habits}} & payment              \\
                                                             & shower time          \\
                                                             & clothes drying       \\
                                                             & broom usage          \\
                                                             & food and drink       \\
                                                             & family               \\
                                                             & popular sports       \\
                                                             & transportation       \\
                                                             & servant              \\
                                                             & wedding              \\
                                                             & funeral       \\
\bottomrule
\end{tabular}
}
\caption{Geo-diverse concept list with categorization.}
\label{tab:list}
\end{table}

We probe GPT-3 with the converted questions in five languages, each of which asks knowledge about the five studied countries. Final results are shown in Table~\ref{gpt3}. One notable result is that using English prompts can achieve nearly 60\% performance, while using Swahili prompts cannot solve any questions correctly. Also for Hindi and Persian prompts, the results are still extremely low, ranging from 0\% to 25\%. It exposes strong bias in terms of language usage. When looking at the performance of probing knowledge about respective countries, the disparity is not large. The country that can be best probed is the United States, while the worst probed country only underperforms the United States 6.9\%.

\section{Detailed Results of Multilingual PLMs on \textsc{GeoMLama}}
\label{results-w}
Table~\ref{xlmfamily},~\ref{mt5family}, and~\ref{xglmfamily} show the details of each multilingual PLM's performance on \textsc{GeoLama}. The performance of random guess depends on the expectation of correct predictions, which is equivalent to the ratio of total number of gold answers to the total number of answers in the answer candidate lists. Since the number of gold answers and answer candidates is different for knowledge about different countries, the random guess performance is not the same across countries. However, prompts in each of the languages have the same number of gold answers and candidate answers, so random guess performance is identical across languages.

\begin{table}[]
\centering
\scalebox{0.73}{
\begin{tabular}{cccccc|c}
\toprule
\textbf{Languages} & \textbf{US} & \textbf{China} & \textbf{India} & \textbf{Iran} & \textbf{Kenya} & \textbf{Average} \\ \midrule
\textbf{en}        & 68.97       & 57.14          & 54.55          & 55.17         & 65.52          & 50.23            \\
\textbf{zh}        & 44.83       & 50.00          & 39.39          & 37.93         & 31.03          & 40.64            \\
\textbf{fa}        & 20.69       & 21.43          & 24.24          & 10.34         & 17.24          & 18.79            \\
\textbf{hi}        & 6.90        & 0.00           & 12.12          & 3.45          & 20.69          & 8.63             \\
\textbf{sw}        & 0.00        & 0.00           & 0.00           & 0.00          & 0.00           & 0.00             \\ \midrule
\textbf{Average}   & 28.28       & 25.71          & 26.06          & 21.38         & 26.90          &  25.67               \\
\bottomrule
\end{tabular}
}
\caption{GPT-3 performance (\%) on \textsc{GeoMLama}.}
\label{gpt3}
\end{table}

\section{Detailed Results of Multilingual PLMs Probed with Prompts without Country Tokens}
\label{results-wo}
Table~\ref{xlmfamily-wo},~\ref{mt5family-wo}, and~\ref{xglmfamily-wo} show the details of each multilingual PLM's performance when input with prompts lacking specified country information. It can help in determining the intrinsic bias of each multilingual PLM.

\begin{table*}[]
\centering
\scalebox{0.8}{
\begin{tabular}{cccccc}
\toprule
\textbf{Languages}                                & \textbf{Countries}                  & \textbf{mBERT} & \textbf{XLM} & \textbf{XLM-R-base} & \textbf{XLM-R-large} \\
\midrule
\multicolumn{1}{c|}{\multirow{5}{*}{\textbf{en}}} & \multicolumn{1}{c|}{\textbf{US}}    & 31.03          & 26.21        & 30.34               & 33.10                 \\
\multicolumn{1}{c|}{}                             & \multicolumn{1}{c|}{\textbf{China}} & 30.00             & 39.29        & 34.29               & 37.14                \\
\multicolumn{1}{c|}{}                             & \multicolumn{1}{c|}{\textbf{India}} & 40.61          & 52.12        & 37.58               & 37.58                \\
\multicolumn{1}{c|}{}                             & \multicolumn{1}{c|}{\textbf{Iran}}  & 21.38          & 27.59        & 28.28               & 37.93                \\
\multicolumn{1}{c|}{}                             & \multicolumn{1}{c|}{\textbf{Kenya}} & 30.63          & 34.38        & 30.63               & 32.50                 \\ \midrule
\multicolumn{1}{c|}{\multirow{5}{*}{\textbf{zh}}} & \multicolumn{1}{c|}{\textbf{US}}    & 35.17          & 28.28        & 30.34               & 46.21                \\
\multicolumn{1}{c|}{}                             & \multicolumn{1}{c|}{\textbf{China}} & 30.71          & 28.57        & 46.43               & 40.00                   \\
\multicolumn{1}{c|}{}                             & \multicolumn{1}{c|}{\textbf{India}} & 38.79          & 32.12        & 38.18               & 35.15                \\
\multicolumn{1}{c|}{}                             & \multicolumn{1}{c|}{\textbf{Iran}}  & 32.41          & 36.55        & 24.14               & 33.10                 \\
\multicolumn{1}{c|}{}                             & \multicolumn{1}{c|}{\textbf{Kenya}} & 41.25          & 35.00           & 27.50                & 39.38                \\ \midrule
\multicolumn{1}{c|}{\multirow{5}{*}{\textbf{fa}}} & \multicolumn{1}{c|}{\textbf{US}}    & 48.97          & 57.93        & 48.28               & 53.79                \\
\multicolumn{1}{c|}{}                             & \multicolumn{1}{c|}{\textbf{China}} & 27.86          & 20.71        & 28.57               & 32.14                \\
\multicolumn{1}{c|}{}                             & \multicolumn{1}{c|}{\textbf{India}} & 38.79          & 27.88        & 33.33               & 34.55                \\
\multicolumn{1}{c|}{}                             & \multicolumn{1}{c|}{\textbf{Iran}}  & 47.59          & 31.03        & 35.17               & 33.79                \\
\multicolumn{1}{c|}{}                             & \multicolumn{1}{c|}{\textbf{Kenya}} & 38.75          & 31.87        & 27.50                & 34.38                \\ \midrule
\multicolumn{1}{c|}{\multirow{5}{*}{\textbf{hi}}} & \multicolumn{1}{c|}{\textbf{US}}    & 42.07          & 40.00           & 33.10               & 42.07                \\
\multicolumn{1}{c|}{}                             & \multicolumn{1}{c|}{\textbf{China}} & 29.29          & 22.86        & 18.57               & 13.57                \\
\multicolumn{1}{c|}{}                             & \multicolumn{1}{c|}{\textbf{India}} & 34.55          & 35.76        & 36.36               & 32.73                \\
\multicolumn{1}{c|}{}                             & \multicolumn{1}{c|}{\textbf{Iran}}  & 33.79          & 31.03        & 31.72               & 27.59                \\
\multicolumn{1}{c|}{}                             & \multicolumn{1}{c|}{\textbf{Kenya}} & 28.75          & 33.75        & 36.25               & 33.75                \\ \midrule
\multicolumn{1}{c|}{\multirow{5}{*}{\textbf{sw}}} & \multicolumn{1}{c|}{\textbf{US}}    & 27.59          & 24.83        & 23.45               & 29.66                \\
\multicolumn{1}{c|}{}                             & \multicolumn{1}{c|}{\textbf{China}} & 34.29          & 22.86        & 32.14               & 29.29                \\
\multicolumn{1}{c|}{}                             & \multicolumn{1}{c|}{\textbf{India}} & 27.88          & 29.70         & 31.52               & 29.09                \\
\multicolumn{1}{c|}{}                             & \multicolumn{1}{c|}{\textbf{Iran}}  & 20.69          & 27.59        & 35.17               & 31.72                \\
\multicolumn{1}{c|}{}                             & \multicolumn{1}{c|}{\textbf{Kenya}} & 26.88          & 31.87        & 27.50                & 31.87                \\
\bottomrule
\end{tabular}
}
\caption{Results (\%) of mBERT, XLM, XLM-R-base, and XLM-R-large on \textsc{GeoMLama}.}
\label{xlmfamily}
\end{table*}

\begin{table*}[]
\centering
\scalebox{0.8}{
\begin{tabular}{ccccc}
\toprule
\textbf{Languages}                                & \textbf{Countries}                  & \textbf{mT5-small} & \textbf{mT5-base} & \textbf{mT5-large} \\ \midrule
\multicolumn{1}{c|}{\multirow{5}{*}{\textbf{en}}} & \multicolumn{1}{c|}{\textbf{US}}    & 24.14              & 18.62             & 30.34              \\
\multicolumn{1}{c|}{}                             & \multicolumn{1}{c|}{\textbf{China}} & 40.71              & 34.29             & 39.29              \\
\multicolumn{1}{c|}{}                             & \multicolumn{1}{c|}{\textbf{India}} & 41.21              & 34.55             & 49.09              \\
\multicolumn{1}{c|}{}                             & \multicolumn{1}{c|}{\textbf{Iran}}  & 19.31              & 19.31             & 26.21              \\
\multicolumn{1}{c|}{}                             & \multicolumn{1}{c|}{\textbf{Kenya}} & 21.88              & 23.75             & 34.38              \\ \midrule
\multicolumn{1}{c|}{\multirow{5}{*}{\textbf{zh}}} & \multicolumn{1}{c|}{\textbf{US}}    & 20.00                 & 33.79             & 28.97              \\
\multicolumn{1}{c|}{}                             & \multicolumn{1}{c|}{\textbf{China}} & 26.43              & 26.43             & 26.43              \\
\multicolumn{1}{c|}{}                             & \multicolumn{1}{c|}{\textbf{India}} & 23.64              & 46.06             & 33.33              \\
\multicolumn{1}{c|}{}                             & \multicolumn{1}{c|}{\textbf{Iran}}  & 33.10               & 26.90              & 31.03              \\
\multicolumn{1}{c|}{}                             & \multicolumn{1}{c|}{\textbf{Kenya}} & 36.88              & 34.38             & 35.00                 \\ \midrule
\multicolumn{1}{c|}{\multirow{5}{*}{\textbf{fa}}} & \multicolumn{1}{c|}{\textbf{US}}    & 55.86              & 43.45             & 48.28              \\
\multicolumn{1}{c|}{}                             & \multicolumn{1}{c|}{\textbf{China}} & 31.43              & 29.29             & 22.86              \\
\multicolumn{1}{c|}{}                             & \multicolumn{1}{c|}{\textbf{India}} & 36.36              & 34.55             & 30.30               \\
\multicolumn{1}{c|}{}                             & \multicolumn{1}{c|}{\textbf{Iran}}  & 28.28              & 30.34             & 33.79              \\
\multicolumn{1}{c|}{}                             & \multicolumn{1}{c|}{\textbf{Kenya}} & 30.00                 & 30.63             & 35.00                 \\ \midrule
\multicolumn{1}{c|}{\multirow{5}{*}{\textbf{hi}}} & \multicolumn{1}{c|}{\textbf{US}}    & 33.79              & 33.79             & 44.14              \\
\multicolumn{1}{c|}{}                             & \multicolumn{1}{c|}{\textbf{China}} & 28.57              & 26.43             & 19.29              \\
\multicolumn{1}{c|}{}                             & \multicolumn{1}{c|}{\textbf{India}} & 33.33              & 33.33             & 35.15              \\
\multicolumn{1}{c|}{}                             & \multicolumn{1}{c|}{\textbf{Iran}}  & 33.79              & 33.10              & 32.41              \\
\multicolumn{1}{c|}{}                             & \multicolumn{1}{c|}{\textbf{Kenya}} & 42.50               & 36.88             & 41.88              \\ \midrule
\multicolumn{1}{c|}{\multirow{5}{*}{\textbf{sw}}} & \multicolumn{1}{c|}{\textbf{US}}    & 37.93              & 32.41             & 28.28              \\
\multicolumn{1}{c|}{}                             & \multicolumn{1}{c|}{\textbf{China}} & 17.86              & 28.57             & 42.86              \\
\multicolumn{1}{c|}{}                             & \multicolumn{1}{c|}{\textbf{India}} & 30.91              & 30.30              & 41.21              \\
\multicolumn{1}{c|}{}                             & \multicolumn{1}{c|}{\textbf{Iran}}  & 36.55              & 26.21             & 23.45              \\
\multicolumn{1}{c|}{}                             & \multicolumn{1}{c|}{\textbf{Kenya}} & 43.12              & 38.75             & 33.75             \\ \bottomrule
\end{tabular}
}
\caption{Results (\%) of models in mT5 family on \textsc{GeoMLama}.}
\label{mt5family}
\end{table*}

\begin{table*}[]
\centering
\scalebox{0.8}{
\begin{tabular}{cccccc}
\toprule
\textbf{Languages}                                & \textbf{Countries}                  & \textbf{XGLM-564M} & \textbf{XGLM-1.7B} & \textbf{XGLM-2.9B} & \textbf{XGLM-4.5B} \\ \midrule
\multicolumn{1}{c|}{\multirow{5}{*}{\textbf{en}}} & \multicolumn{1}{c|}{\textbf{US}}    & 32.41              & 37.93              & 31.72              & 37.24              \\
\multicolumn{1}{c|}{}                             & \multicolumn{1}{c|}{\textbf{China}} & 37.86              & 32.14              & 39.29              & 35.71              \\
\multicolumn{1}{c|}{}                             & \multicolumn{1}{c|}{\textbf{India}} & 30.91              & 40.00                 & 43.03              & 42.42              \\
\multicolumn{1}{c|}{}                             & \multicolumn{1}{c|}{\textbf{Iran}}  & 23.45              & 28.28              & 20.00                 & 31.03              \\
\multicolumn{1}{c|}{}                             & \multicolumn{1}{c|}{\textbf{Kenya}} & 21.88              & 25.00                 & 26.25              & 35.00                 \\ \midrule
\multicolumn{1}{c|}{\multirow{5}{*}{\textbf{zh}}} & \multicolumn{1}{c|}{\textbf{US}}    & 34.48              & 36.55              & 40.00                 & 35.86              \\
\multicolumn{1}{c|}{}                             & \multicolumn{1}{c|}{\textbf{China}} & 25.71              & 33.57              & 30.00                 & 37.14              \\
\multicolumn{1}{c|}{}                             & \multicolumn{1}{c|}{\textbf{India}} & 27.27              & 32.73              & 36.36              & 31.52              \\
\multicolumn{1}{c|}{}                             & \multicolumn{1}{c|}{\textbf{Iran}}  & 18.62              & 22.07              & 25.52              & 13.79              \\
\multicolumn{1}{c|}{}                             & \multicolumn{1}{c|}{\textbf{Kenya}} & 24.38              & 19.38              & 16.88              & 20.00                 \\ \midrule
\multicolumn{1}{c|}{\multirow{5}{*}{\textbf{fa}}} & \multicolumn{1}{c|}{\textbf{US}}    & 49.66              & 49.66              & 46.90               & 49.66              \\
\multicolumn{1}{c|}{}                             & \multicolumn{1}{c|}{\textbf{China}} & 26.43              & 27.86              & 25.71              & 35.00                 \\
\multicolumn{1}{c|}{}                             & \multicolumn{1}{c|}{\textbf{India}} & 32.73              & 31.52              & 28.48              & 32.73              \\
\multicolumn{1}{c|}{}                             & \multicolumn{1}{c|}{\textbf{Iran}}  & 37.24              & 35.86              & 31.72              & 36.55              \\
\multicolumn{1}{c|}{}                             & \multicolumn{1}{c|}{\textbf{Kenya}} & 34.38              & 30.00                 & 33.75              & 27.50               \\ \midrule
\multicolumn{1}{c|}{\multirow{5}{*}{\textbf{hi}}} & \multicolumn{1}{c|}{\textbf{US}}    & 35.86              & 28.97              & 33.79              & 28.97              \\
\multicolumn{1}{c|}{}                             & \multicolumn{1}{c|}{\textbf{China}} & 18.57              & 10.00                 & 20.71              & 21.43              \\
\multicolumn{1}{c|}{}                             & \multicolumn{1}{c|}{\textbf{India}} & 33.33              & 29.70               & 29.09              & 33.94              \\
\multicolumn{1}{c|}{}                             & \multicolumn{1}{c|}{\textbf{Iran}}  & 34.48              & 22.76              & 32.41              & 23.45              \\
\multicolumn{1}{c|}{}                             & \multicolumn{1}{c|}{\textbf{Kenya}} & 34.38              & 26.88              & 30.00                 & 26.25              \\ \midrule
\multicolumn{1}{c|}{\multirow{5}{*}{\textbf{sw}}} & \multicolumn{1}{c|}{\textbf{US}}    & 25.52              & 28.28              & 27.59              & 24.14              \\
\multicolumn{1}{c|}{}                             & \multicolumn{1}{c|}{\textbf{China}} & 33.57              & 38.57              & 30.71              & 30.71              \\
\multicolumn{1}{c|}{}                             & \multicolumn{1}{c|}{\textbf{India}} & 34.55              & 34.55              & 37.58              & 33.33              \\
\multicolumn{1}{c|}{}                             & \multicolumn{1}{c|}{\textbf{Iran}}  & 31.72              & 22.07              & 21.38              & 33.10               \\
\multicolumn{1}{c|}{}                             & \multicolumn{1}{c|}{\textbf{Kenya}} & 26.88              & 29.38              & 30.63              & 33.75             \\ \bottomrule
\end{tabular}
}
\caption{Results (\%) of models in XGLM family on \textsc{GeoMLama}.}
\label{xglmfamily}
\end{table*}

\begin{table*}[]
\centering
\scalebox{0.8}{
\begin{tabular}{cccccc}
\toprule
\textbf{Languages}                                & \textbf{Countries}                  & \textbf{mBERT} & \textbf{XLM} & \textbf{XLM-R-base} & \textbf{XLM-R-large} \\ \midrule
\multicolumn{1}{c|}{\multirow{5}{*}{\textbf{en}}} & \multicolumn{1}{c|}{\textbf{US}}    & 31.03          & 26.21        & 30.34               & 33.10                 \\
\multicolumn{1}{c|}{}                             & \multicolumn{1}{c|}{\textbf{China}} & 30.00             & 39.29        & 34.29               & 37.14                \\
\multicolumn{1}{c|}{}                             & \multicolumn{1}{c|}{\textbf{India}} & 40.61          & 52.12        & 37.58               & 37.58                \\
\multicolumn{1}{c|}{}                             & \multicolumn{1}{c|}{\textbf{Iran}}  & 21.38          & 27.59        & 28.28               & 37.93                \\
\multicolumn{1}{c|}{}                             & \multicolumn{1}{c|}{\textbf{Kenya}} & 30.63          & 34.38        & 30.63               & 32.50                 \\ \midrule
\multicolumn{1}{c|}{\multirow{5}{*}{\textbf{zh}}} & \multicolumn{1}{c|}{\textbf{US}}    & 35.17          & 28.28        & 30.34               & 46.21                \\
\multicolumn{1}{c|}{}                             & \multicolumn{1}{c|}{\textbf{China}} & 30.71          & 28.57        & 46.43               & 40.00                   \\
\multicolumn{1}{c|}{}                             & \multicolumn{1}{c|}{\textbf{India}} & 38.79          & 32.12        & 38.18               & 35.15                \\
\multicolumn{1}{c|}{}                             & \multicolumn{1}{c|}{\textbf{Iran}}  & 32.41          & 36.55        & 24.14               & 33.10                 \\
\multicolumn{1}{c|}{}                             & \multicolumn{1}{c|}{\textbf{Kenya}} & 41.25          & 35.00           & 27.50                & 39.38                \\ \midrule
\multicolumn{1}{c|}{\multirow{5}{*}{\textbf{fa}}} & \multicolumn{1}{c|}{\textbf{US}}    & 48.97          & 57.93        & 48.28               & 53.79                \\
\multicolumn{1}{c|}{}                             & \multicolumn{1}{c|}{\textbf{China}} & 27.86          & 20.71        & 28.57               & 32.14                \\
\multicolumn{1}{c|}{}                             & \multicolumn{1}{c|}{\textbf{India}} & 38.79          & 27.88        & 33.33               & 34.55                \\
\multicolumn{1}{c|}{}                             & \multicolumn{1}{c|}{\textbf{Iran}}  & 47.59          & 31.03        & 35.17               & 33.79                \\
\multicolumn{1}{c|}{}                             & \multicolumn{1}{c|}{\textbf{Kenya}} & 38.75          & 31.87        & 27.50                & 34.38                \\ \midrule
\multicolumn{1}{c|}{\multirow{5}{*}{\textbf{hi}}} & \multicolumn{1}{c|}{\textbf{US}}    & 42.07          & 40.00           & 33.10                & 42.07                \\
\multicolumn{1}{c|}{}                             & \multicolumn{1}{c|}{\textbf{China}} & 29.29          & 22.86        & 18.57               & 13.57                \\
\multicolumn{1}{c|}{}                             & \multicolumn{1}{c|}{\textbf{India}} & 34.55          & 35.76        & 36.36               & 32.73                \\
\multicolumn{1}{c|}{}                             & \multicolumn{1}{c|}{\textbf{Iran}}  & 33.79          & 31.03        & 31.72               & 27.59                \\
\multicolumn{1}{c|}{}                             & \multicolumn{1}{c|}{\textbf{Kenya}} & 28.75          & 33.75        & 36.25               & 33.75                \\ \midrule
\multicolumn{1}{c|}{\multirow{5}{*}{\textbf{sw}}} & \multicolumn{1}{c|}{\textbf{US}}    & 27.59          & 24.83        & 23.45               & 29.66                \\
\multicolumn{1}{c|}{}                             & \multicolumn{1}{c|}{\textbf{China}} & 34.29          & 22.86        & 32.14               & 29.29                \\
\multicolumn{1}{c|}{}                             & \multicolumn{1}{c|}{\textbf{India}} & 27.88          & 29.70         & 31.52               & 29.09                \\
\multicolumn{1}{c|}{}                             & \multicolumn{1}{c|}{\textbf{Iran}}  & 20.69          & 27.59        & 35.17               & 31.72                \\
\multicolumn{1}{c|}{}                             & \multicolumn{1}{c|}{\textbf{Kenya}} & 26.88          & 31.87        & 27.50                & 31.87               \\ \bottomrule
\end{tabular}
}
\caption{Results (\%) of mBERT, XLM, XLM-R-base, XLM-R-large probed
with prompts without country tokens
on \textsc{GeoMLama}.}
\label{xlmfamily-wo}
\end{table*}

\begin{table*}[]
\centering
\scalebox{0.8}{
\begin{tabular}{ccccc}
\toprule
\textbf{Languages}                                & \textbf{Countries}                  & \textbf{mT5-small} & \textbf{mT5-base} & \textbf{mT5-large} \\ \midrule
\multicolumn{1}{c|}{\multirow{5}{*}{\textbf{en}}} & \multicolumn{1}{c|}{\textbf{US}}    & 38.62              & 49.66             & 40.69              \\
\multicolumn{1}{c|}{}                             & \multicolumn{1}{c|}{\textbf{China}} & 45.00                 & 47.14             & 42.86              \\
\multicolumn{1}{c|}{}                             & \multicolumn{1}{c|}{\textbf{India}} & 46.06              & 51.52             & 60.61              \\
\multicolumn{1}{c|}{}                             & \multicolumn{1}{c|}{\textbf{Iran}}  & 38.62              & 46.90              & 43.45              \\
\multicolumn{1}{c|}{}                             & \multicolumn{1}{c|}{\textbf{Kenya}} & 43.12              & 44.38             & 57.50               \\ \midrule
\multicolumn{1}{c|}{\multirow{5}{*}{\textbf{zh}}} & \multicolumn{1}{c|}{\textbf{US}}    & 24.14              & 24.83             & 30.34              \\
\multicolumn{1}{c|}{}                             & \multicolumn{1}{c|}{\textbf{China}} & 32.86              & 30.71             & 33.57              \\
\multicolumn{1}{c|}{}                             & \multicolumn{1}{c|}{\textbf{India}} & 35.15              & 28.48             & 31.52              \\
\multicolumn{1}{c|}{}                             & \multicolumn{1}{c|}{\textbf{Iran}}  & 36.55              & 40.69             & 40.00                 \\
\multicolumn{1}{c|}{}                             & \multicolumn{1}{c|}{\textbf{Kenya}} & 39.38              & 43.12             & 36.88              \\ \midrule
\multicolumn{1}{c|}{\multirow{5}{*}{\textbf{fa}}} & \multicolumn{1}{c|}{\textbf{US}}    & 46.21              & 41.38             & 47.59              \\
\multicolumn{1}{c|}{}                             & \multicolumn{1}{c|}{\textbf{China}} & 39.29              & 33.57             & 41.43              \\
\multicolumn{1}{c|}{}                             & \multicolumn{1}{c|}{\textbf{India}} & 34.55              & 41.82             & 35.76              \\
\multicolumn{1}{c|}{}                             & \multicolumn{1}{c|}{\textbf{Iran}}  & 35.86              & 37.24             & 42.76              \\
\multicolumn{1}{c|}{}                             & \multicolumn{1}{c|}{\textbf{Kenya}} & 37.50               & 41.88             & 42.50               \\ \midrule
\multicolumn{1}{c|}{\multirow{5}{*}{\textbf{hi}}} & \multicolumn{1}{c|}{\textbf{US}}    & 31.72              & 25.52             & 29.66              \\
\multicolumn{1}{c|}{}                             & \multicolumn{1}{c|}{\textbf{China}} & 39.29              & 38.57             & 32.86              \\
\multicolumn{1}{c|}{}                             & \multicolumn{1}{c|}{\textbf{India}} & 44.24              & 46.67             & 41.21              \\
\multicolumn{1}{c|}{}                             & \multicolumn{1}{c|}{\textbf{Iran}}  & 30.34              & 33.79             & 31.03              \\
\multicolumn{1}{c|}{}                             & \multicolumn{1}{c|}{\textbf{Kenya}} & 40.00                 & 40.00                & 41.25              \\ \midrule
\multicolumn{1}{c|}{\multirow{5}{*}{\textbf{sw}}} & \multicolumn{1}{c|}{\textbf{US}}    & 22.76              & 28.28             & 25.52              \\
\multicolumn{1}{c|}{}                             & \multicolumn{1}{c|}{\textbf{China}} & 23.57              & 35.71             & 42.86              \\
\multicolumn{1}{c|}{}                             & \multicolumn{1}{c|}{\textbf{India}} & 30.91              & 35.76             & 37.58              \\
\multicolumn{1}{c|}{}                             & \multicolumn{1}{c|}{\textbf{Iran}}  & 14.48              & 20.69             & 22.07              \\
\multicolumn{1}{c|}{}                             & \multicolumn{1}{c|}{\textbf{Kenya}} & 16.88              & 23.75             & 26.25             \\ \bottomrule
\end{tabular}
}
\caption{Results (\%) of models in mT5 family probed with prompts without country tokens
on \textsc{GeoMLama}.}
\label{mt5family-wo}
\end{table*}

\begin{table*}[]
\centering
\scalebox{0.8}{
\begin{tabular}{cccccc}
\toprule
\textbf{Languages}                                & \textbf{Countries}                  & \textbf{XGLM-564M} & \textbf{XGLM-1.7B} & \textbf{XGLM-2.9B} & \textbf{XGLM-4.5B} \\ \midrule
\multicolumn{1}{c|}{\multirow{5}{*}{\textbf{en}}} & \multicolumn{1}{c|}{\textbf{US}}    & 28.97              & 38.62              & 34.48              & 40.00                 \\
\multicolumn{1}{c|}{}                             & \multicolumn{1}{c|}{\textbf{China}} & 57.14              & 43.57              & 50.00                 & 46.43              \\
\multicolumn{1}{c|}{}                             & \multicolumn{1}{c|}{\textbf{India}} & 51.52              & 47.88              & 53.94              & 46.67              \\
\multicolumn{1}{c|}{}                             & \multicolumn{1}{c|}{\textbf{Iran}}  & 35.86              & 35.17              & 34.48              & 36.55              \\
\multicolumn{1}{c|}{}                             & \multicolumn{1}{c|}{\textbf{Kenya}} & 40.62              & 49.38              & 43.75              & 46.88              \\ \midrule
\multicolumn{1}{c|}{\multirow{5}{*}{\textbf{zh}}} & \multicolumn{1}{c|}{\textbf{US}}    & 34.48              & 42.76              & 38.62              & 47.59              \\
\multicolumn{1}{c|}{}                             & \multicolumn{1}{c|}{\textbf{China}} & 49.29              & 55.00                 & 51.43              & 50.71              \\
\multicolumn{1}{c|}{}                             & \multicolumn{1}{c|}{\textbf{India}} & 44.24              & 52.73              & 54.55              & 46.67              \\
\multicolumn{1}{c|}{}                             & \multicolumn{1}{c|}{\textbf{Iran}}  & 54.48              & 52.41              & 46.21              & 63.45              \\
\multicolumn{1}{c|}{}                             & \multicolumn{1}{c|}{\textbf{Kenya}} & 55.62              & 58.13              & 62.50               & 61.25              \\ \midrule
\multicolumn{1}{c|}{\multirow{5}{*}{\textbf{fa}}} & \multicolumn{1}{c|}{\textbf{US}}    & 27.59              & 28.97              & 35.17              & 34.48              \\
\multicolumn{1}{c|}{}                             & \multicolumn{1}{c|}{\textbf{China}} & 34.29              & 37.86              & 35.00                 & 40.00                 \\
\multicolumn{1}{c|}{}                             & \multicolumn{1}{c|}{\textbf{India}} & 38.18              & 34.55              & 40.00                & 36.97              \\
\multicolumn{1}{c|}{}                             & \multicolumn{1}{c|}{\textbf{Iran}}  & 17.93              & 22.07              & 24.83              & 24.14              \\
\multicolumn{1}{c|}{}                             & \multicolumn{1}{c|}{\textbf{Kenya}} & 21.88              & 28.12              & 30.63              & 33.12              \\ \midrule
\multicolumn{1}{c|}{\multirow{5}{*}{\textbf{hi}}} & \multicolumn{1}{c|}{\textbf{US}}    & 24.14              & 32.41              & 20.69              & 31.72              \\
\multicolumn{1}{c|}{}                             & \multicolumn{1}{c|}{\textbf{China}} & 52.86              & 52.86              & 48.57              & 55.71              \\
\multicolumn{1}{c|}{}                             & \multicolumn{1}{c|}{\textbf{India}} & 39.39              & 41.21              & 40.61              & 40.61              \\
\multicolumn{1}{c|}{}                             & \multicolumn{1}{c|}{\textbf{Iran}}  & 37.93              & 39.31              & 28.28              & 42.07              \\
\multicolumn{1}{c|}{}                             & \multicolumn{1}{c|}{\textbf{Kenya}} & 36.25              & 41.25              & 41.88              & 43.12              \\ \midrule
\multicolumn{1}{c|}{\multirow{5}{*}{\textbf{sw}}} & \multicolumn{1}{c|}{\textbf{US}}    & 42.07              & 40.00                 & 41.38              & 35.17              \\
\multicolumn{1}{c|}{}                             & \multicolumn{1}{c|}{\textbf{China}} & 42.14              & 39.29              & 36.43              & 27.86              \\
\multicolumn{1}{c|}{}                             & \multicolumn{1}{c|}{\textbf{India}} & 40.00                 & 42.42              & 50.30               & 46.06              \\
\multicolumn{1}{c|}{}                             & \multicolumn{1}{c|}{\textbf{Iran}}  & 33.10               & 24.83              & 37.24              & 31.72              \\
\multicolumn{1}{c|}{}                             & \multicolumn{1}{c|}{\textbf{Kenya}} & 42.50               & 41.25              & 46.25              & 32.50              \\ \bottomrule
\end{tabular}
}
\caption{Results (\%) of models in XGLM family probed with prompts without country tokens
on \textsc{GeoMLama}.}
\label{xglmfamily-wo}
\end{table*}

\end{document}